# Low-dose Spectral CT Reconstruction Using Image Gradient $\ell_0$–norm and Tensor Dictionary

**Weiwen Wu**[1,2], **Yanbo Zhang**[2], **Qian Wang**[2], **Fenglin Liu**[1,3,*], **Peijun Chen**[1] **and Hengyong Yu**[2]
[1]Key Lab of Optoelectronic Technology and Systems, Ministry of Education, Chongqing University, Chongqing 400044, China
[2]Department of Electrical and Computer Engineering, University of Massachusetts Lowell, Lowell, MA 01854, USA
[3]Engineering Research Center of Industrial Computed Tomography Nondestructive Testing, Ministry of Education, Chongqing University, Chongqing 400044, China
*Corresponding author: Fenglin Liu
E-mail: liufl@cqu.edu.cn.

**Abstract:** Spectral computed tomography (CT) has a great superiority in lesion detection, tissue characterization and material decomposition. To further extend its potential clinical applications, in this work, we propose an improved tensor dictionary learning method for low-dose spectral CT reconstruction with a constraint of image gradient $\ell_0$-norm, which is named as $\ell_0$TDL. The $\ell_0$TDL method inherits the advantages of tensor dictionary learning (TDL) by employing the similarity of spectral CT images. On the other hand, by introducing the $\ell_0$-norm constraint in gradient image domain, the proposed method emphasizes the spatial sparsity to overcome the weakness of TDL on preserving edge information. The split-bregman method is employed to solve the proposed method. Both numerical simulations and real mouse studies are perform to evaluate the proposed method. The results show that the proposed $\ell_0$TDL method outperforms other competing methods, such as total variation (TV) minimization, TV with low rank (TV+LR), and TDL methods.

**Keywords:** spectral computed tomography (CT), image reconstruction, low-dose, sparse-view, $\ell_0$-norm of image gradient, tensor dictionary.

## 1. Introduction

As an imaging tool, x-ray computed tomography (CT) has been widely applied in clinical diagnosis, industrial detection and security inspections [1, 2]. However, there are some inherent weaknesses in the conventional CT. First，the traditional CT images have no sufficient tissue contrast resolution and material decomposition analysis capability [3]. Second, the dose is relatively high which implies high risks especially for children [4]. Fortunately, the multi-energy CT (spectroscopic, spectral or color CT) has attracted continuous attentions for its superior performance in term of material decomposition, tissue characterization and lesion detection [5, 6].

Dual energy CT (DECT), as the simplest spectral CT, has been used in many applications, such as material decomposition [7], abdomen angiography detection [8], and pulmonary artery sarcoma and pulmonary embolism identification [9]. However, the DECT has two main limitations. First, the DECT still utilizes the energy-integrating detectors and results in spectral overlap. Second, the DECT only acquires two attenuation intensities at either two different x-ray source spectra or two energy windows. Thus, only limited material decomposition maps can be discriminated from the dual energy CT data [10].

Different from the DECT, the state-of-the-art spectral CT scanner (e.g. MARS [11]) employs photon-counting detectors (PCDs) to record the energy of each individually incoming x-ray photon by converting the electronic pulse signal of the quanta to the corresponding peak amplitudes of photon energy [12]. Thus, we can obtain the material decomposition maps from multiple projection datasets in different energy bins after a series of post-processing steps. Indeed, the spectral CT has achieved tremendous successes in low-dose CT [13], contrast media imaging [14], and K-edge imaging [15]. However, many physical effects can introduce errors in the number and energy of photons measured by the PCD [16]. One non-ideal effect is that photons are recorded in incorrect energy channels due to cross-talk. Another non-ideal effect is the overlap of energy window which means photons with energy outside of the window thresholds can contribute to the energy window measurement. All of these can result in noisy measured dataset by the PCD. This leads to lower signal noise ratio (SNR) measurements and compromises the material decomposition results[17]. As pointed out by the MARS's team [11], it is a primary challenge to generate accurate and clean volumetric material images for spectral CT. To further extend the clinical applications of spectral CT (breast spectral CT [18], solitary pulmonary nodules spectral CT [19], *etc.*), in this work we will address the low-dose spectral CT reconstruction issue.

To obtain high quality reconstructed spectral CT images from noisy projection datasets, a number of image reconstruction methods have been reported. Elbakri *et al.* constructed a penalized-likelihood function for the multi-energetic model and further developed an ordered-subsets iterative method to estimate the unknown material for each voxel [20]. Xu *et al.* considered each channel spectral projection data as an independent traditional dataset and applied the total variation (TV) penalty to reconstruct interior ROI spectral images [21]. To suppress the disturbance of global intensity and protect the image edge, a PRISM (prior rank, intensity, and sparsity model) technique was utilized for multi-energy CT reconstruction [22, 23]. Xu *et al.* developed dictionary learning methods for the conventional low-dose CT image reconstruction and spectral CT reconstruction [24]. To achieve high-quality spectral breast CT images from few-view projections, a tight-frame based iterative reconstruction (TFIR) technique was investigated in [25]. Sawatzky *et al.* explored a multi-channel penalized weighted least squares (PWLS) estimator to improve image quality [26]. Rigie and La Rivière incorporated a total nuclear variation ($TV_N$) into the Chambolle and Pock primal-dual algorithm for spectral CT [27]. Because a small patch often contains only one or two materials, a patch-based low-rank penalty was proposed for sparse-view kVp switching-based spectral CT [5]. Xi *et al.* designed two types of united iterative reconstruction (UIR) algorithms to characterize the structure correlations of images in the energy domain [28]. To suppress noise within a narrower energy bin in spectral imaging, a high-quality spectral mean image as prior information was introduced into the prior image constrained compressed sensing (PICCS) algorithm [29] and then generated spectral PICCS [30]. To exploit the correlations among all the dimensions simultaneously and edge-preserving/enhancement, Semerci *et al.* combined a tensor nuclear norm (TNN) with TV for multi-energy reconstruction [31].

The tensor dictionary learning (TDL) was derived from the conventional vectorized dictionary [32] learning by extending vector-matrix to higher tensor data for obtaining better image classification results [33]. A decomposable nonlocal tensor dictionary learning (DNTDL) further considered the non-local similarity of tensor patch for multispectral image (MSI) denoising [34]. An orthogonal tensor dictionary learning was developed for dynamic texture recognition [35]. Considering the image similarity of reconstructed images among different time frames, the TDL was also applied to 4D CT reconstruction [36]. Recently, a TDL method was developed by considering the similarity of spectral CT images from different energy channels [37]. Although such TDL for spectral CT reconstruction algorithm can obtain a relatively better performance in preserving fine structures, it is not good at preserving edge information.

As a regularization term, the image gradient $\ell_0$–norm minimization was introduced in image smoothing [38] and then was applied to image segmentation [39], sparse linear hyperspectral unmixing [40], sparse blind deconvolution [41, 42], image restoration [43] and breast tissue classification [44]. To maintain the inherent image edges and reduce limited-angle artifacts, the $\ell_0$-norm of image gradient was introduced into limited-angle CT reconstruction [45]. Very recently, Yu *et al*. proposed an iterative reconstruction method based on the image gradient $\ell_0$–norm minimization to recover the image from incomplete datasets smeared by sparse-view and limited-angle artifacts [46]. The main advantage is the image gradient $\ell_0$-norm penalizes the number of non-zero image gradient rather than the image gradient magnitudes. As a result, the proposed method is less sensitive to the intensity changes, and it can reserve edge directions and recover fine structures. The image gradient $\ell_0$-norm also emphasizes image spatial sparsity, which can reduce artifacts and improve the ability of denoising for the proposed algorithm.

To overcome the aforementioned limitations of the TDL method, we combine the image gradient $\ell_0$-norm minimization and the TDL technique to generate an $\ell_0$TDL algorithm. The contributions of this work are mainly threefold. First, to suppress the staircase artifacts and overcome the parameter selection issue derived from the TV constraint, we incorporate the image gradient $\ell_0$-norm into the functional of the TDL method to generate the $\ell_0$TDL algorithm. Second, an efficient alternating direction minimization method (ADMM) is developed for the proposed $\ell_0$TDL model for low-dose spectral CT iterative reconstruction. Third, the $\ell_0$TDL parameters are optimized with extensive experiments. The proposed $\ell_0$TDL method has the following advantages: i) introducing the image gradient $\ell_0$-norm to encourage the TDL-based method to recover fine structures and edge information; ii) improving the image sparsity to further suppress noise and reduce image artifacts.

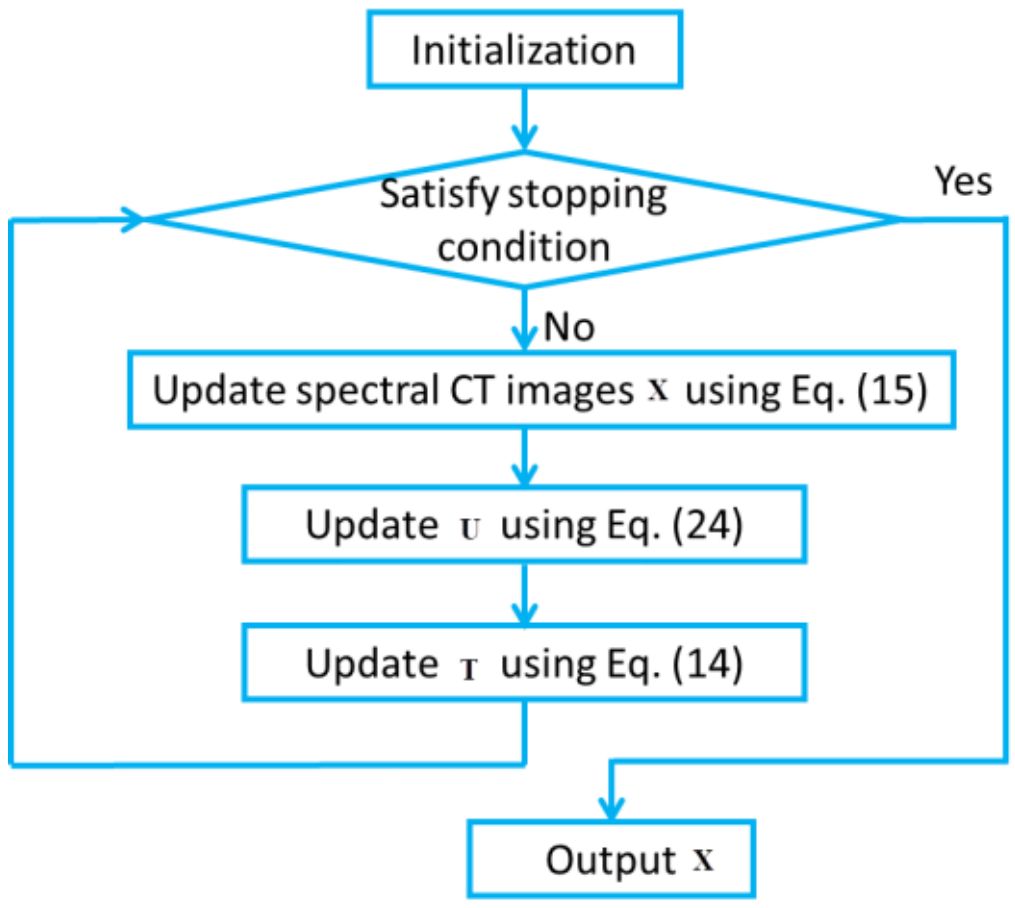

Fig. 1. The flowchart of the $\ell_0$TDL method.

The rest of this paper is organized as follows. In section II, we briefly review the traditional TDL algorithm and image gradient $\ell_0$-norm minimization. In section III, we establish the $\ell_0$TDL mathematic model and the corresponding split-bregman solution, and the parameters for $\ell_0$TDL method are also analyzed. In section IV, both numerical simulations and preclinical mouse studies are performed to evaluate the developed algorithm. In section V, we discuss some related issues and conclude the paper.

## 2. Reconstruction method

### *2.A. Tensor dictionary learning*

A tensor is a multidimensional data array. A N$^{th}$ order tensor can be defined as $\mathbf{X} \in \Re^{I_1 \times I_2 \times \ldots \times I_N}$, whose element is $x_{i_1 \times i_2 \times \ldots i_N}$, $1 \le i_n \le I_n$ and $n = 1, 2, \ldots, N$. Particularly, if $N$ equals 1 or 2, the corresponding tensor would be degraded into a vector or matrix. A tensor can also be multiplied by a vector or a matrix. Therefore, the mode-$n$ product of a tensor $\mathbf{x}$ by a matrix $\mathbf{H} \in \Re^{J \times I_n}$ can be defined by $\mathbf{X} \times_n \mathbf{H} \in \Re^{I_1 \times I_2 \times \ldots \times I_{n-1} \times J \times I_{n+1} \times \ldots \times I_N}$, whose element in $\Re^{I_1 \times I_2 \times \ldots \times I_{n-1} \times J \times I_{n+1} \times \ldots \times I_N}$ can be calculated by $\sum_{i_n=1}^{I_n} x_{i_1 \times i_2 \times \ldots i_N} h_{j \times i_n}$. In this work, we only consider the case that $\mathbf{X}$ is a 3$^{rd}$ tensor.

Suppose that there are a set of 3$^{rd}$-order tensors $\mathbf{X}^{(t)} \in \Re^{I_1 \times I_2 \times I_3}$ and $t = 1, 2, \ldots, T$. The tensor-based dictionary learning can be converted to solve the following optimization problem:

$$\underset{\boldsymbol{D},\boldsymbol{\alpha_t}}{\operatorname{argmin}} \sum_{t=1}^{T} \left\| \mathbf{X}^{(t)} - \mathbf{D} \times_4 \boldsymbol{\alpha}_t \right\|_F^2 \quad s.t. \ \left\| \boldsymbol{\alpha}_t \right\|_0 \le L, \tag{1}$$

where $\mathbf{D} = \left\{ \mathbf{D}^{(k)} \right\} \in \Re^{I_1 \times I_2 \times I_3 \times K}$ is the tensor dictionary, and $K$ and $L$ represent the number of atoms in the dictionary and the sparsity level, respectively. $\|\cdot\|_F$ and $\|\cdot\|_0$ represent the Frobenius- and $\ell_0$-norm, respectively. The K-CPD algorithm can be employed to train a tensor dictionary [47]. The solution of objective function (1) can be obtained by using the alternative direction minimization method (ADMM). The first step is to update the sparse coefficient matrix by using the multilinear orthogonal matching pursuit (MOMP) technique and fixing the tensor dictionary $\mathbf{D}$ [47]. Then, the second step is to update the tensor dictionary with fixed sparse coefficient matrix. With the alternative update procedures, the desired tensor dictionary and the corresponding sparse representation coefficients can be achieved simultaneously.

### *2.B.* Tensor dictionary learning for spectral CT

The TDL model for 2D spectral CT reconstruction can be expressed as follow [37],

$$\underset{\mathbf{X},\boldsymbol{\alpha}_r,\mathbf{m}_r}{\arg\min} \sum_{s=1}^{S} \left\| \mathbf{A}\mathbf{x}_s - \mathbf{y}_s \right\|_2^2 + \lambda \left( \sum_r \left\| \mathbb{Z}_r(\mathbf{X}) - \mathbf{D}_m \times_4 \mathbf{m}_r - \mathbf{D} \times_4 \boldsymbol{\alpha}_r \right\|_F^2 + \sum_r \kappa_r \left\| \boldsymbol{\alpha}_r \right\|_0 \right). \tag{2}$$

where $\mathbf{X} \in \Re^{I_1 \times I_2 \times S}$ and $\mathbf{Y} \in \Re^{J_1 \times J_2 \times S}$ are respectively the 3$^{rd}$-order reconstructed image and projection data tensors, $I_1$ and $I_2$ are width and height of the reconstructed image, $J_1$ and $J_2$ present the number of detector and projection views, $S$ is a number of energy channels, $\mathbf{x}_s$ and $\mathbf{y}_s$ are respectively the vectorized $s^{th}$ image and projection, $\mathbf{A}$ is the system matrix which depended on the system scanning structure and calculation method, $\mathbf{m}_r$ presents the mean vector of each channel, the operator $\mathbb{Z}_r$ is used to extract $r^{th}$ small tensor block ($N \times N \times S$) from $\mathbf{X}$ and $\boldsymbol{\alpha}_r \in \Re^K$ is the sparse representation coefficient of $r^{th}$ tensor block. The $\mathbf{D} = \left\{ \mathbf{D}^{(k)} \right\} \in \Re^{N \times N \times S \times K}$ is the trained tensor dictionary and $\mathbf{D}_m = \left\{ \mathbf{D}_m^{(k)} \right\} \in \Re^{N \times N \times S \times S}$ represents the mean removal process [48]. The parameters of $\kappa_r$ is a factor to modulate representation precision and the sparsity level, and $\lambda$ is designed to balance the data fidelity term and the sparse representation regularization. The system matrix $\mathbf{A}$ from specified scanning configuration has a large impact on the parameter $\lambda$. To make parameter $\lambda$ more stable, it can be expressed as follow

$$\lambda = \frac{\eta S \sum_{i_1=1}^{I_1} \sum_{i_2=1}^{I_2} \left[\mathbf{A}^{\mathrm{T}}\mathbf{A}\right]_{i_1 i_2}}{\sum_r \sum_{s=1}^{S} \sum_{i_1=1}^{I_1} \sum_{i_2=1}^{I_2} \left[\mathbb{Z}_r{}^{\mathrm{T}}\mathbb{Z}_r\right]_{i_1 i_2}}, \tag{3}$$

where the symbols $[\cdot]_{i_1 i_2}$ and $[\cdot]_{i_1 i_2 s}$ represent the $(i_1, i_2)^{\text{th}}$ element of a matrix and $(i_1, i_2, s)^{\text{th}}$ element of a given tensor, respectively. The parameter $\eta$ is a scaled parameter to balance the data fidelity term and sparse representation regularization term. To solve the optimization problem (2), an alternating minimization strategy for $\mathbf{X}$, $\mathbf{m}_r$ and $\boldsymbol{\alpha}_r$ was proposed in [37] and the method can be divided into three steps. The first step aims to minimize $\mathbf{X}$ by updating $\mathbf{X}$ using Eq. (4)

$$\mathbf{X}_{i_1 i_2 s}^{n+1} = \mathbf{X}_{i_1 i_2 s}^{n} - \frac{\left[\mathbf{A}^{\mathrm{T}}\left(\mathbf{A}\mathbf{x}_s^n - \mathbf{y}_s\right)\right]_{i_1 i_2} + \lambda \sum_r \mathbb{Z}_r^{\mathrm{T}}\left(\mathbb{Z}_r\left(\mathbf{X}^n\right) - \mathbf{D}_m \times_4 \mathbf{m}_r^n - \mathbf{D}\times_4 \boldsymbol{\alpha}_r^n\right)_{i_1 i_2 s}}{\left[\mathbf{A}^{\mathrm{T}}\mathbf{A}\right]_{i_1 i_2} + \lambda\left[\sum_r \mathbb{Z}_r^{\mathrm{T}}\mathbb{Z}_r\right]_{i_1 i_2 s}}, \tag{4}$$

where the operator $\mathbb{Z}_r^{\mathrm{T}}$ rearranges a tensor patch into the reconstructed image tensor space. Then,

$\mathbf{m}_r^{n+1}$ can be updated by calculating the following sub-problem

$$\mathbf{m}_r^{n+1} = \arg\min_{\mathbf{m}_r} \left\| \mathbb{Z}_r\left(\mathbf{X}^{n+1}\right) - \mathbf{D}_m \times_4 \mathbf{m}_r - \mathbf{D}\times_4 \boldsymbol{\alpha}_r^n \right\|_F^2 . \tag{5}$$

In the last step, the sparse representation matrix $\boldsymbol{\alpha}_r^{n+1}$ can be updated as Eq. (6)

$$\boldsymbol{\alpha}_r^{n+1} = \arg\min_{\boldsymbol{\alpha}_r} \left\| \mathbb{Z}_r\left(\mathbf{X}^{n+1}\right) - \mathbf{D}_m \times_4 \mathbf{m}_r^{n+1} - \mathbf{D}\times_4 \boldsymbol{\alpha}_r \right\|_F^2 + \kappa_r \left\|\boldsymbol{\alpha}_r\right\|_0 . \tag{6}$$

### *2.C.* Image gradient $\ell_0$-norm minimization

Image gradient $\ell_0$-norm, different from the natural image $\ell_0$-norm, has been proposed to enhance image smoothing [38] and extended to image segmentation, visual enhancement, *etc*. It also has been applied to the limited-angle and sparse angle problems, resulting high-quality CT images [45, 46]. The $\ell_0$-norm of image gradient can be denoted as

$$\left\|\nabla \mathbf{x}_s\right\|_0 = \sum_p \#\left\{p \,\middle|\, \left|\partial_x \mathbf{x}_s^p\right| + \left|\partial_y \mathbf{x}_s^p\right| \neq 0\right\}, \tag{7}$$

where $\#\{\cdot\}$ is a counting operator and $p$ ($p$=1, 2, …, $I_1 \times I_2$) index the location of $(i_1, i_2)^{\text{th}}$ element in the image. The $\partial_x \mathbf{x}_s^p$ and $\partial_y \mathbf{x}_s^p$ represent $\left(\mathbf{x}_s(i_1,i_2) - \mathbf{x}_s(i_1-1,i_2)\right)$ and $\left(\mathbf{x}_s(i_1,i_2) - \mathbf{x}_s(i_1,i_2-1)\right)$, respectively. If $\left|\mathbf{x}_s(i_1,i_2) - \mathbf{x}_s(i_1-1,i_2)\right| + \left|\mathbf{x}_s(i_1,i_2) - \mathbf{x}_s(i_1,i_2-1)\right| \neq 0$, the counting operator would add 1. Eq. (7) shows that the gradient magnitude is not considered by the $\ell_0$-norm. That is to say, greater gradient magnitudes are not penalized by the image gradient $\ell_0$-norm which results in an effective preservation of edge information and fine structures.

## 3. $\ell_0$TDL reconstruction method

### *3.A.* $\ell_0$TDL mathematic model

Because the image edges and fine structures are corrupted by severe noise in low-dose spectral CT reconstruction, the TDL may fail to recover high-quality edge information from such an undersampling dataset. To achieve a better image with more accurate edge information and less noise suppression, it is natural for us to combine the image gradient $\ell_0$–norm minimization and TDL technology. As a result, we formulate an image reconstruction framework as follow:

$$\arg\min_{\mathbf{X},\boldsymbol{\alpha}_r,\mathbf{m}_r} \sum_{s=1}^{S} \|\mathbf{A}\mathbf{x}_s - \mathbf{y}_s\|_2^2 + \mu \sum_{s=1}^{S} \|\nabla \mathbf{x}_s\|_0 + \lambda \left( \sum_r \|\mathbb{Z}_r(\mathbf{X}) - \mathbf{D}_m \times_4 \mathbf{m}_r - \mathbf{D} \times_4 \boldsymbol{\alpha}_r\|_F^2 + \sum_r \kappa_r \|\boldsymbol{\alpha}_r\|_0 \right). \quad (8)$$

## 3.B. Solution

Because Eq. (8) contains three searched-for variables, we further divide it into three sub-problems:

$$\mathbf{X}^{n+1} = \arg\min_{\mathbf{X}} \sum_{s=1}^{S} \|\mathbf{A}\mathbf{x}_s - \mathbf{y}_s\|_2^2 + \mu \sum_{s=1}^{S} \|\nabla \mathbf{x}_s\|_0 + \lambda \sum_r \|\mathbb{Z}_r(\mathbf{X}) - \mathbf{D}_m \times_4 \mathbf{m}_r^n - \mathbf{D} \times_4 \boldsymbol{\alpha}_r^n\|_F^2 , \quad (9a)$$

$$\mathbf{m}_r^{n+1} = \arg\min_{\mathbf{m}_r} \|\mathbb{Z}_r(\mathbf{X}^{n+1}) - \mathbf{D}_m \times_4 \mathbf{m}_r - \mathbf{D} \times_4 \boldsymbol{\alpha}_r^n\|_F^2 , \quad (9b)$$

$$\boldsymbol{\alpha}_r^{n+1} = \arg\min_{\boldsymbol{\alpha}_r} \|\mathbb{Z}_r(\mathbf{X}^{n+1}) - \mathbf{D}_m \times_4 \mathbf{m}_r^{n+1} - \mathbf{D} \times_4 \boldsymbol{\alpha}_r\|_F^2 + \kappa_r \|\boldsymbol{\alpha}_r\|_0 . \quad (9c)$$

Eqs. (9b) and (9c) can be easily solved by following the same steps in [37]. Eq. (9a) contains the $\ell_0$-norm of image gradient and tensor dictionary based sparse representation, which is non-convex and non-deterministic polynomial hard (NP-hard) problem. To solve this optimization problem effectively, we employ an alternating direction minimization method (ADMM). First, we introduce an auxiliary variable $\mathbf{u}_S$. Eq. (9a) can be re-expressed as a constrained optimization model:

$$\arg\min_{\mathbf{X}} \sum_{s=1}^{S} \|\mathbf{A}\mathbf{x}_s - \mathbf{y}_s\|_2^2 + \lambda \sum_r \|\mathbb{Z}_r(\mathbf{X}) - \mathbf{D}_m \times_4 \mathbf{m}_r^n - \mathbf{D} \times_4 \boldsymbol{\alpha}_r^n\|_F^2 + \mu \sum_{s=1}^{S} \|\nabla \mathbf{u}_s\|_0 , \text{ s.t. } \mathbf{u}_s = \mathbf{x}_s , \quad (10)$$

where $\mathbf{u}_S$ is an auxiliary matrix in $\Re^{I_1 \times I_2}$ for the $s^{\text{th}}$ energy channel which is an element of tensor $\mathbf{U}$ in $\Re^{I_1 \times I_2 \times S}$. Thus, the scaled augmented Lagrangian function of problem (10) [49] can be written as

$$\arg\min_{\mathbf{X},\mathbf{U},\mathbf{T}} \sum_{s=1}^{S} \|\mathbf{A}\mathbf{x}_s - \mathbf{y}_s\|_2^2 + \lambda \sum_r \|\mathbb{Z}_r(\mathbf{X}) - \mathbf{D}_m \times_4 \mathbf{m}_r^n - \mathbf{D} \times_4 \boldsymbol{\alpha}_r^n\|_F^2 + \mu \sum_{s=1}^{S} \|\nabla \mathbf{u}_s\|_0 + \beta \sum_{s=1}^{S} \|\mathbf{x}_s - \mathbf{u}_s - \mathbf{t}_s\|_F^2 , \quad (11)$$

where $\mathbf{t}_s$ is an auxiliary variable in $\Re^{I_1 \times I_2}$ for the $s^{\text{th}}$ energy channel which is a cell of tensor $\mathbf{T}$ in $\Re^{I_1 \times I_2 \times S}$. In fact, the ADMM method is utilized to iteratively and alternately solve Eq. (11) with respect to $\mathbf{X}$, $\mathbf{U}$ and $\mathbf{T}$. $\beta$ is the Lagrangian multiplier for all energy channel. The split-bregman algorithm of Eq. (11) contains the following three steps:

$$\mathbf{X}^{n+1} = \arg\min_{\mathbf{X}} \sum_{s=1}^{S} \|\mathbf{A}\mathbf{x}_s - \mathbf{y}_s\|_2^2 + \lambda \sum_r \|\mathbb{Z}_r(\mathbf{X}) - \mathbf{D}_m \times_4 \mathbf{m}_r^n - \mathbf{D} \times_4 \boldsymbol{\alpha}_r^n\|_F^2 + \beta \sum_{s=1}^{S} \|\mathbf{x}_s - \mathbf{u}_s^n - \mathbf{t}_s^n\|_F^2 , \quad (12)$$

$$\mathbf{U}^{n+1} = \arg\min_{\mathbf{U}} \mu \sum_{s=1}^{S} \|\nabla \mathbf{u}_s\|_0 + \beta \sum_{s=1}^{S} \|\mathbf{x}_s^{n+1} - \mathbf{u}_s - \mathbf{t}_s^n\|_F^2 , \quad (13)$$

$$\mathbf{T}^{n+1} = \mathbf{T}^n + \mathbf{U}^{n+1} - \mathbf{X}^{n+1}. \quad (14)$$

In this work, the solution of Eq. (12) can be given by the separable surrogate method and the form can be expressed as follow:

$$\mathbf{X}_{i_1 i_2 s}^{n+1} = \mathbf{X}_{i_1 i_2 s}^{n} - \frac{\left[\mathbf{A}^{\mathrm{T}}\left(\mathbf{A}\mathbf{x}_s^n - \mathbf{y}_s\right)\right]_{i_1 i_2} + \lambda \sum_r \mathbb{Z}_r^{\mathrm{T}}\left(\mathbb{Z}_r\left(\mathbf{X}^n\right) - \mathbf{D}_m \times_4 \mathbf{m}_r^n - \mathbf{D} \times_4 \boldsymbol{\alpha}_r^n\right)_{i_1 i_2 s} + \beta\left[\left(\mathbf{x}_s^n - \mathbf{u}_s^n - \mathbf{t}_s^n\right)\right]_{i_1 i_2}}{\left[\mathbf{A}^{\mathrm{T}}\mathbf{A}\right]_{i_1 i_2} + \lambda\left[\sum_r \mathbb{Z}_r^{\mathrm{T}}\mathbb{Z}_r\right]_{i_1 i_2 s} + \beta} . \quad (15)$$

Eq. (13) includes the $\ell_0$-norm minimization of image gradient, resulting in a non-convex and NP-hard problem. Fortunately, an approximate method was proposed in [38] to solve this problem.

For the approximate method, another two auxiliary variables $\left(h_s^p, v_s^p\right)$ corresponding to the gradients $\left(\partial_x \mathbf{u}_s^p, \partial_y \mathbf{u}_s^p\right)$ are introduced. Therefore, Eq. (13) can be converted into the following problem

$$\mathbf{u}_s^{n+1} = \underset{\mathbf{u}_s, h_s, v_s}{\arg\min} \sum_p \left\| \left(\mathbf{x}_s^p\right)^{n+1} - \mathbf{u}_s^p - \left(\mathbf{t}_s^p\right)^n \right\|_F^2 + \lambda^* \Gamma\left(\mathbf{h}_s, \mathbf{v}_s\right) + \tau\left(\left(\partial x\left(\mathbf{u}_s^p\right) - h_s^p\right)^2 + \left(\partial y\left(\mathbf{u}_s^p\right) - v_s^p\right)^2\right), \quad (16)$$

where $\lambda^* = \mu / \beta$, $\Gamma(h_s, v_s) = \left\{p \left\| h_s^p \right| + \left| v_s^p \right| \neq 0\right\}$, $\left(\mathbf{x}_s^p\right)^{n+1}$ and $\left(\mathbf{t}_s^p\right)^n$ are components of the $p^{th}$ pixel in ($n$+1) and $n$ iterations, and $\tau$ aims to balance the similarity between $\left(\mathbf{h}_s, \mathbf{v}_s\right)$ and $\left(\partial_x u_s^p, \partial_y u_s^p\right)$. $\Gamma\left(\mathbf{h}_s, \mathbf{v}_s\right)$ can be expressed as:

$$\Gamma\left(\mathbf{h}_s, \mathbf{v}_s\right) = \sum_p \Lambda\left(\left|h_s^p\right| + \left|v_s^p\right|\right), \quad (17)$$

where

$$\Lambda\left(\left|h_s^p\right| + \left|v_s^p\right|\right) = \begin{cases} 1 & if \ \left|h_s^p\right| + \left|v_s^p\right| \neq 0 \\ 0 & \text{otherwise} \end{cases}. \quad (18)$$

Because Eq. (16) contains three variables, we also employ the split-bregman scheme, *i.e.* updating one or multiple variables and fixing others. Thus, the optimization problem of Eq. (16) can be divided into the following two steps:

$$\left\{\mathbf{h}_s^{m+1}, \mathbf{v}_s^{m+1}\right\} = \underset{\mathbf{h}_s, \mathbf{v}_s}{\arg\min} \sum_p \left(\left(\partial_x \left(\mathbf{u}_s^p\right)^n - h_s^p\right)^2 + \left(\partial_y \left(\mathbf{u}_s^p\right)^n - v_s^p\right)^2\right) + \frac{\lambda^*}{\tau} \Gamma\left(\mathbf{h}_s, \mathbf{v}_s\right), \quad (19)$$

$$\mathbf{u}_s^{n+1} = \underset{\mathbf{u}_s}{\arg\min} \sum_p \left\| \left(\mathbf{x}_s^p\right)^{n+1} - \mathbf{u}_s^p + \left(\mathbf{t}_s^p\right)^n \right\|_F^2 + \tau\left(\left(\partial_x \left(\mathbf{u}_s^p\right) - \left(h_s^p\right)^{m+1}\right)^2 + \left(\partial_y \left(\mathbf{u}_s^p\right) - \left(v_s^p\right)^{m+1}\right)^2\right). \quad (20)$$

Substituting Eq. (17) into (19), we have:

$$\left\{\mathbf{h}_s^{m+1}, \mathbf{v}_s^{m+1}\right\} = \underset{\mathbf{h}_s, \mathbf{v}_s}{\arg\min} \sum_p \left(\left(\partial_x \left(\mathbf{u}_s^p\right)^n - h_s^p\right)^2 + \left(\partial_y \left(\mathbf{u}_s^p\right)^n - v_s^p\right)^2 + \frac{\lambda^*}{\tau} \Lambda\left(\left|h_s^p\right| + \left|v_s^p\right|\right)\right). \quad (21)$$

Because each pixel $\mathbf{u}_s^p$ is considered independent in the iterative process of (21), we can separate each pixel so that it can be addressed independently. Thus, Eq. (21) can be rewritten as

$$\left\{\left(h_s^p\right)^{m+1}, \left(v_s^p\right)^{m+1}\right\} = \underset{\mathbf{h}_s, \mathbf{v}_s}{\arg\min} \sum_p \left(\left(\partial_x \left(\mathbf{u}_s^p\right)^n - h_s\right)^2 + \left(\partial_y \left(\mathbf{u}_s^p\right)^n - v_s\right)^2 + \frac{\lambda^*}{\tau} \Lambda\left(\left|h_s\right| + \left|v_s\right|\right)\right). \quad (22)$$

For Eq. (22), the energy function can easily reach its minimum with the optimization criteria as follow:

$$\left(\left(h_s^p\right)^{m+1}, \left(v_s^p\right)^{m+1}\right) = \begin{cases} 1 & \left(\partial_x \left(\mathbf{u}_s^p\right)^n\right)^2 + \left(\partial_y \left(\mathbf{u}_s^p\right)^n\right)^2 \leq \frac{\lambda^*}{\tau} \\ 0 & otherwise \end{cases}. \quad (23)$$

Now, let's consider Eq. (20). Because the function is quadratic, even a gradient descent method can make it shrink to a global minimum solution. Alternatively, we employ a fast analytic technique [46, 50] which integrates diagonalization derivative operators after Fast Fourier Transform (FFT). Therefore, the solution of formula (20) reads,

$$\mathbf{u}_s^{n+1} = \mathbb{F}^{-1}\left\{ \frac{\mathbb{F}\left(\mathbf{x}_s^{n+1} - \mathbf{t}_s^{n}\right) + \tau\left(\left(\mathbb{F}^*\left(\partial_x\right)\right)\left(\mathbb{F}^* h_s^{m+1}\right) + \left(\mathbb{F}^*\partial_y\right)\left(\mathbb{F}^* v_s^{m+1}\right)\right)}{\mathbb{F}(1) + \tau\left(\left(\mathbb{F}^*\left(\partial_x\right)\right)\left(\mathbb{F}^*\partial_x\right) + \left(\mathbb{F}^*\left(\partial_y\right)\right)\left(\mathbb{F}^*\partial_y\right)\right)} \right\} . \quad (24)$$

Where $\mathbb{F}$ and $\mathbb{F}^*$ represent Fourier transform and conjugate Fourier transform respectively. The Eq. (13) can be solved by employing the gradient image $\ell_0$–norm minimization algorithm. In summary, the corresponding pseudo code is presented in algorithm I.

**Algorithm I: Image gradient $\ell_0$-norm minimization**

**Input**: $\mathbf{W} \leftarrow \mathbf{X}^{n+1} - \mathbf{T}^{n}$ , $m \leftarrow 0$ , $\mathbf{U}^{n} \leftarrow 0$ , $\lambda^*$ , $\tau^{(0)} = 2\lambda^*$ , $\tau_{\max} = 10^5$ ;
**Output:** $\mathbf{U}^{n+1}$ ;
**While** ( $\tau \le \tau_{\max}$ )
**do**
**For** *s=1:S*
Updating $\left\{\left(h_s^p\right)^{m+1}, \left(v_s^p\right)^{m+1}\right\}$ using Eq. (23);
Updating $\mathbf{u}_s^{m+1}$ by employing Eq. (24);
**end**
$\tau \leftarrow 1.1 \times \tau$ , $m \leftarrow m+1$ ;
**end while**
$\mathbf{U}^{n+1} \leftarrow \mathbf{U}^{m}$
Returning the intermediate result $\mathbf{U}^{n+1}$

In order to unify the $\ell_0$-norm gradient term parameter $\beta$ and the parameter $\lambda$ of tensor dictionary term, the parameter $\beta$ can be given as follow,

$$\beta = \frac{\sigma S \sum_{i_1=1}^{I_1} \sum_{i_2=1}^{I_2} \left[\mathbf{A}^{\mathrm{T}}\mathbf{A}\right]_{i_1 i_2}}{\sum_r \sum_{s=1}^{S} \sum_{i_1=1}^{I_1} \sum_{i_2=1}^{I_2} \left[\mathbb{Z}_r^{\mathrm{T}} \mathbb{Z}_r\right]_{i_1 i_2}}, \quad (29)$$

where $\sigma$ is a scaled parameter to balance the $\ell_0$-norm gradient term and the data fidelity term and dictionary learning term. To perform the proposed $\ell_0$TDL algorithm for low-dose spectral reconstruction, we summarize the main workflow as algorithm II.

**Algorithm II: $\ell_0$TDL**

**Input**:
TDL parameters: $\eta$ , $\varepsilon$ , $K := 1024$ , $L$ and other parameters.
$\ell_0$–norm minimization parameters: $\sigma$ , $\lambda^*$ , $\tau_0 \leftarrow 2\lambda^*$ , $\tau_{\max} \leftarrow 10^5$ .
Initialization of $\mathbf{X}^{(0)}$ , $\mathbf{U} \leftarrow 0$ , $\mathbf{T} \leftarrow 0$ ;
**Output:** reconstructed low-dose image $\mathbf{X}$
**Part I: Dictionary training**
Normalizing the full projection datasets;
Obtaining reconstruct images utilizing FBP from full projection noisy datasets or real datasets;
Extracting patches and training a global tensor dictionary $\mathbf{D}$ using the K-CPD.
**Part II: low-dose image reconstruction**
**While** the stopping criteria are not satisfied
**do**
Updating $\mathbf{X}^{n+1}$ by Eq. (15);
Updating $\mathbf{U}^{n+1}$ utilizing algorithm I;
Updating $\mathbf{T}^{n+1}$ by Eq. (14);
Updating $\mathbf{m}^{n+1}$ based on Eq. (5);
Updating $\boldsymbol{\alpha}^{n+1}$ by adopting the MOMP algorithm;
Positive constraint on $\mathbf{X}^{n+1}$ ;
**end while**
Denormalizing the reconstructed image.
Returning the final result $\mathbf{X}$

## *3.C.* Selection of parameters

The minimization problem in Eq. (8) includes two regularization terms which need a number of parameters for optimization. First, the TDL term mainly includes three parameters: sparse level $L$, precision level ε and the number of atoms $K$. In fact, $K$ can be fixed by satisfying $K > N\times N\times S$ ($N$ is the patch size) and $K$ is set as 1024 [37] in this work. Second, the image gradient $\ell_0$–norm based optimization framework mainly depends on the smoothness control factor $\lambda^*$ in this study [46]. To balance the functions of two regularization programs, the regularization parameters $\sigma$ and $\eta$ are introduced. Different choices of these parameters may lead to different reconstructed images. To study the performance of the $\ell_0$TDL algorithm with respect to different parameters, we only relax one or two free parameters while other parameters are fixed. To quantitatively evaluate the performance of different parameters for the proposed algorithms, the indexes of root means square error (RMSE), feature similarity (FSIM) and structural similarity (SSIM) are employed on parameter selection, reconstructed channel images and decomposed material images.

*1) Regularization parameters $\sigma$ and $\eta$*: In this study, we explore the influence of different regularization parameters on the image quality by extensive experiments. Fig. 2 shows that the differences are small with respect to different $\sigma$ in term of RMSE. The greater the parameter $\sigma$ is, the higher the SSIM and FSIM are, especially for high energy channels. However, a greater $\sigma$ can make the reconstructed image blur and further lose finer structures (see Fig. 3). Because a great $\sigma$ value is set, the composition of image gradient $\ell_0$-norm results in an over-smoothing image. Therefore, it is very important to make a trade-off between $\sigma$ and image quality based the specified reconstruction requirements. The parameter $\eta$ controls the proposition of TDL in the model. The stronger relationship among different energy channels is, the bigger parameter $\eta$ is. To demonstrate the effect of $\eta$ on the image quality, we set $\eta$ as a series of values. Fig. 3 presents the image quality with respect to different $\eta$ values. From Figs. 3 and 4, we can infer the parameter $\eta$ may have little impact on image quality when it is compared with $\sigma$. It is necessary to emphasize that the regularization parameters $\sigma$ and $\eta$ are dependent each other. Again, if the image gradient $\ell_0$–norm part is fixed, we can reach an optimization solution by empirically adjusting the parameter $\eta$.

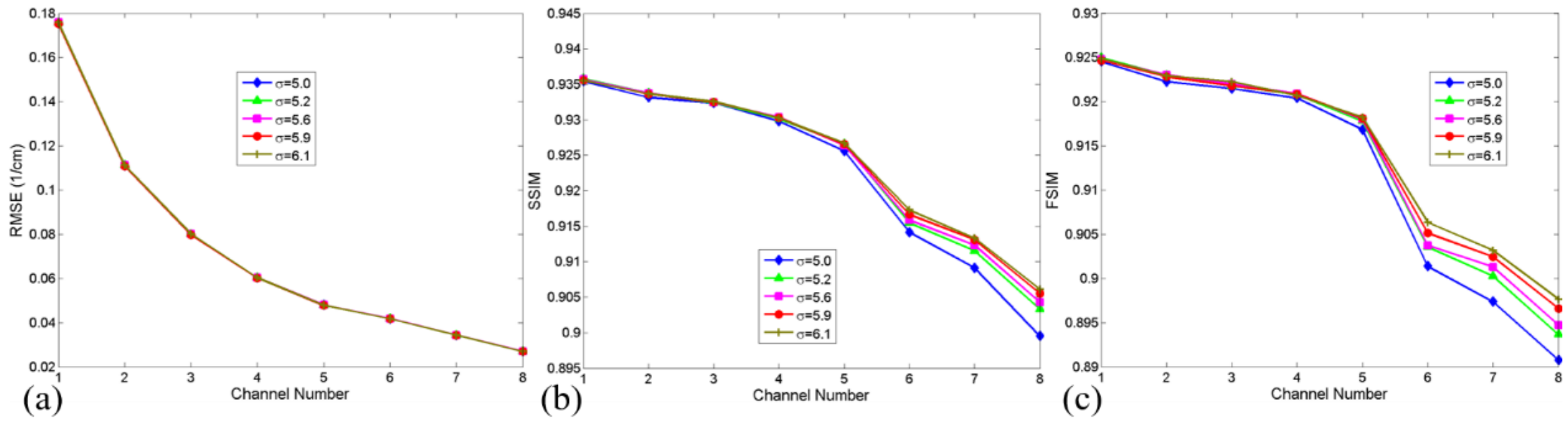

Fig. 2. Image quality assessments of the reconstructed images for the $\ell_0$TDL method with respect to different $\sigma$.

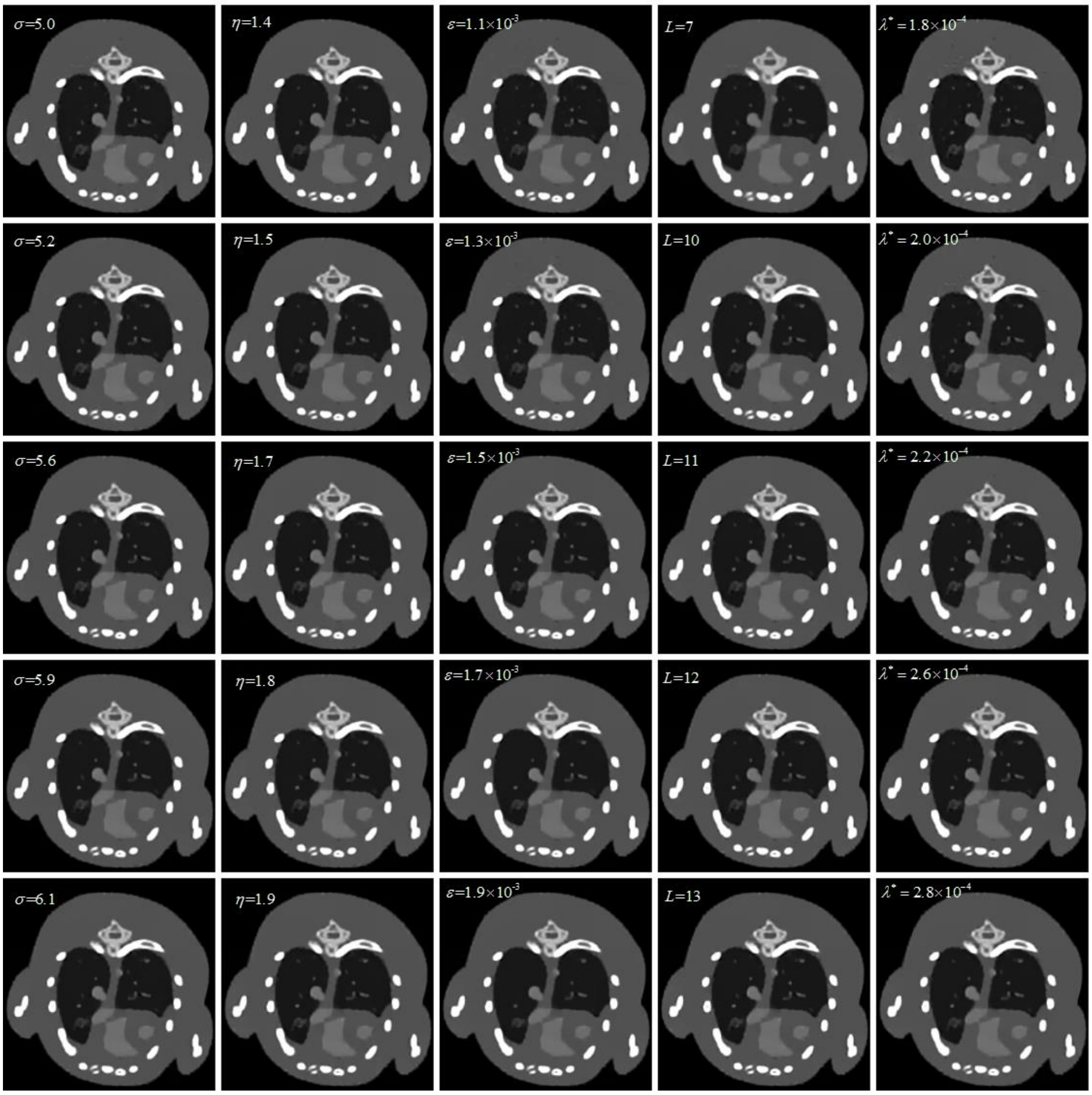

Fig. 3. Representative image slice of the mouse thorax phantom (channel 1) reconstructed by the $\ell_0$TDL algorithm with different parameter settings. Each column represents different values of the same parameter and the rest parameters are fixed. The display window is [0 3] cm$^{-1}$.

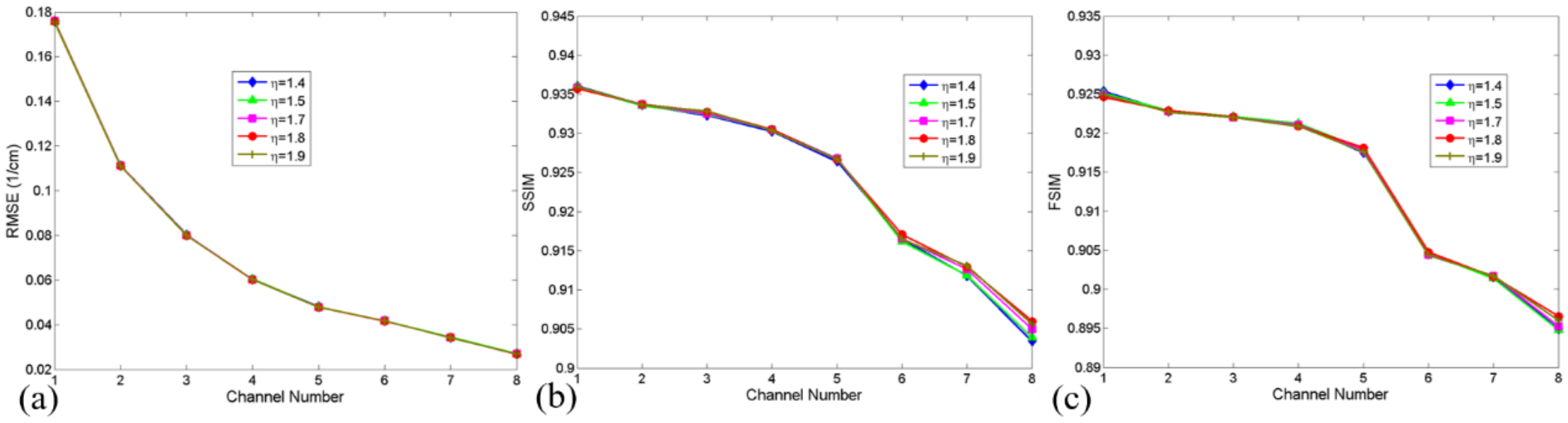

Fig. 4. Same as Fig.2 but for different $\eta$ .

2) *TDL parameters $\varepsilon$ and L:* It is observed from Figs. 3 and 5 that the parameter $\varepsilon$ plays an important role in controlling the final reconstructed images. A greater $\varepsilon$ can improve the image quality to some extent, but it will also induce the loss of finer structures. A smaller $\varepsilon$ may have poor performance in anti-noising and results in some salt and pepper noises, especially for high-energy channels. To investigate the effect of the sparsity level $L$ on image quality, a series of different $L$ values are selected and the assessment results are displayed in Fig. 6. From Figs. 3 and 6, we can observe that a lower sparsity level may results in better results in terms of protecting the image edge-information. Contrary to general expectation, a higher sparsity level may lose finer structure information.

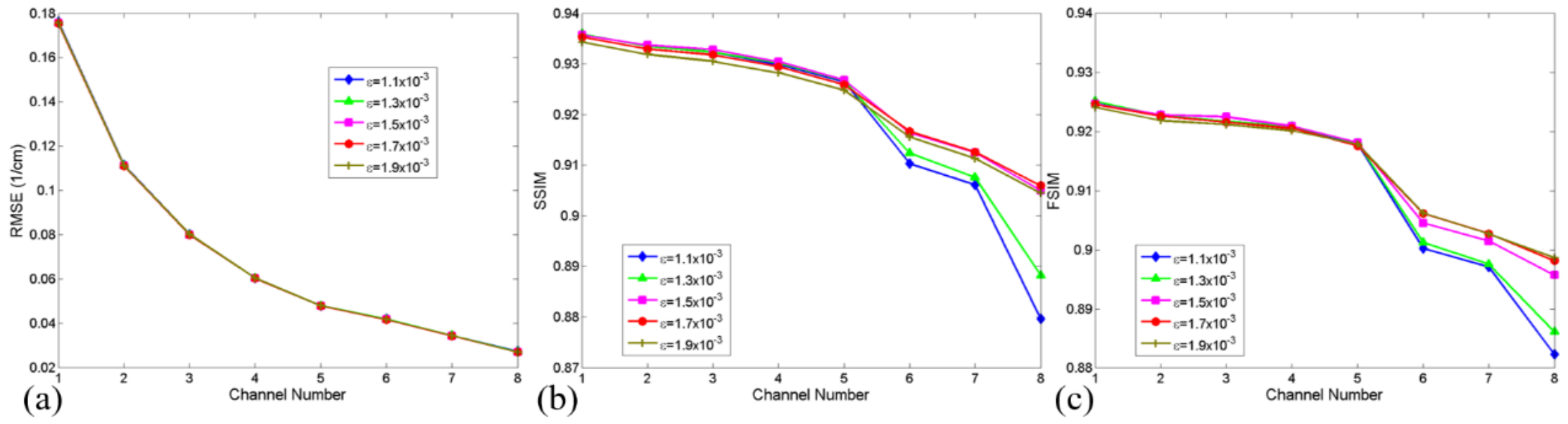

Fig. 4. Same as Fig.1 but for different $\varepsilon$ .

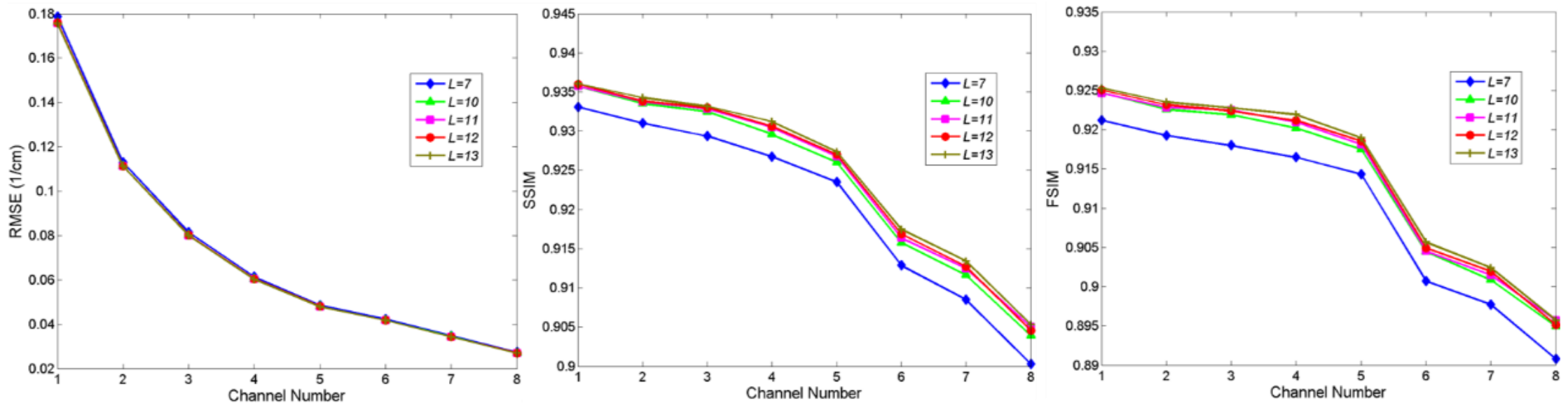

Fig. 6. Same as Fig. 2 but for different $L$.

*3) Smooth factor* $\lambda^*$*:* The parameter $\lambda^*$ is designed to control the smoothness of the reconstructed image. A greater $\lambda^*$ not only improves the image quality but also protects the image edge, which is validated by the experiment results shown in Figs. 3 and 7. According to Eq. (23), it is easy for us to understand that the parameter $\lambda^*$ can not be continuously increased without any limitation. Otherwise, the reconstructed image will be oversmoothing and the edge information will be lost.

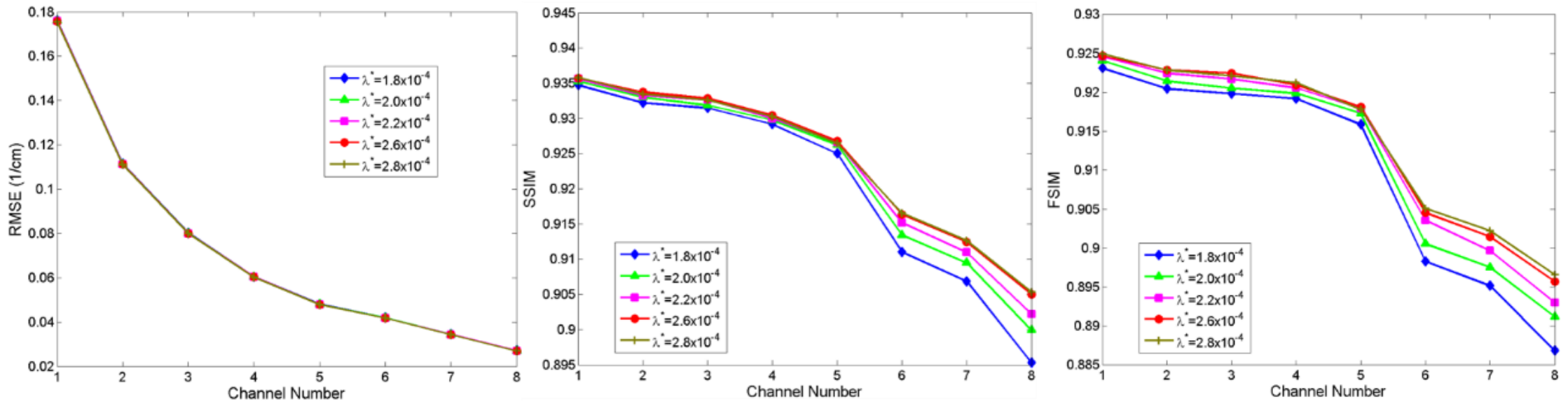

Fig. 7. Same as Fig.2 but for different $\lambda^*$.

# 4. Experiment Results

In this study, we employ both numerical simulations and real spectral data to validate and evaluate the proposed algorithm. The traditional filtered back-projection (FBP), TV minimization, TV+LR and TDL algorithms are implemented and chosen as comparisons. The FBP and TV were done on all energy channels one-by-one. All the aforementioned approaches are carried out on a PC (16 CPUs @3.40GHz, 32.0GB RAM) in Matlab (2017b). For all the iterative methods, the initial images are set as FBP reconstruction, the iteration number is 200 for numerical simulations and 100 for real experiments. Moreover, we employ the ordered subset SART (OS-SART) technique [51] to accelerate the convergence, where the subset number is fixed to 10. For the TDL and $\ell_0$TDL methods, the full-projection-based FBP reconstruction is employed for global tensor dictionary training. Finally, the optimized parameters of the proposed method are listed in table 1 for reconstruction.

Table 1. Image reconstruction parameters for both numerical simulation and realistic dataset.

| | Photon Number | Views | σ | η | ε | λ* | L |
|---|---|---|---|---|---|---|---|
| Numerical simulation | $5\times10^3$ | 160 | 4.80 | 1.10 | $1.10\times10^{-3}$ | $1.80\times10^{-4}$ | 13 |
| | | 106 | 5.30 | 1.40 | $1.25\times10^{-3}$ | $2.45\times10^{-4}$ | 12 |
| | | 80 | 5.70 | 1.60 | $1.50\times10^{-3}$ | $2.60\times10^{-4}$ | 11 |
| | $4\times10^3$ | 80 | 5.80 | 1.60 | $1.60\times10^{-3}$ | $2.60\times10^{-4}$ | 11 |
| | $3\times10^3$ | 80 | 6.10 | 1.90 | $2.10\times10^{-3}$ | $3.10\times10^{-4}$ | 9 |
| Realistic dataset | | 120 | 3.20 | 1.10 | $7.00\times10^{-4}$ | $6.50\times10^{-5}$ | 12 |
| | | 80 | 5.00 | 1.40 | $7.00\times10^{-4}$ | $8.00\times10^{-5}$ | 11 |
| | | 40 | 5.40 | 1.60 | $9.00\times10^{-4}$ | $1.20\times10^{-4}$ | 10 |

## *4.A. Numerical simulations*

### *4.A.1 Sparse-view image reconstruction*

In the numerical simulations, we employ a simulated mouse thorax phantom, where 1.2% iodine contrast agent is injected to the blood circulation (Fig. 8) [52]. A 50KVp x-ray spectrum is utilized, and it is divided into 8 energy channels: [16, 22) keV, [22, 25) keV, [25, 28) keV, [28, 31) keV, [31, 34) keV, [34, 37) keV, [37, 41) keV, and [41, 50) keV [37]. The geometrical protocol is as follows: the distances from x-ray source to the PCD and rotation center are 180*mm* and 132*mm,* respectively; the PCD consists of 512 units and each of which is 0.1*mm*; the size of reconstructed channel image is 256×256×8 and each pixel covers an area of 0.15×0.15mm². We collect 640 projections for each full scan, and the default photon number is 5000 for each x-ray beam. For all the simulations, Poisson noises are superimposed on the obtained projections.

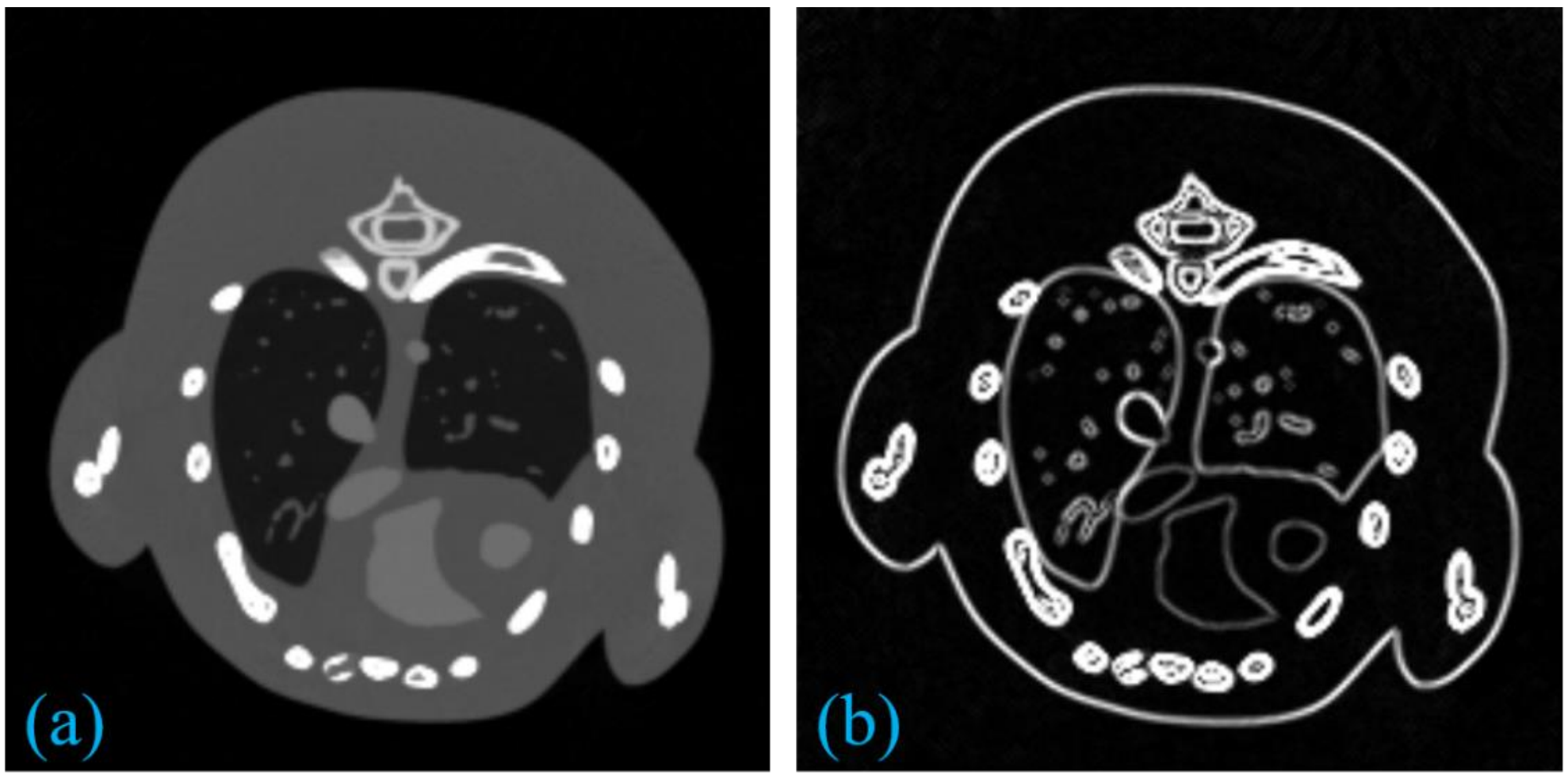

Fig 8. The mouse thorax phantom (left) and the corresponding gradient image (right).

To demonstrate the advantages of our proposed algorithm in recovering high-quality images from sparse-view projections, reconstructed images from 160, 106 and 80 views are shown in Figs.

9, 10 and 11, along with the counterparts from other competing algorithms. For simplicity, we only show the reconstructed images for the first energy channel.

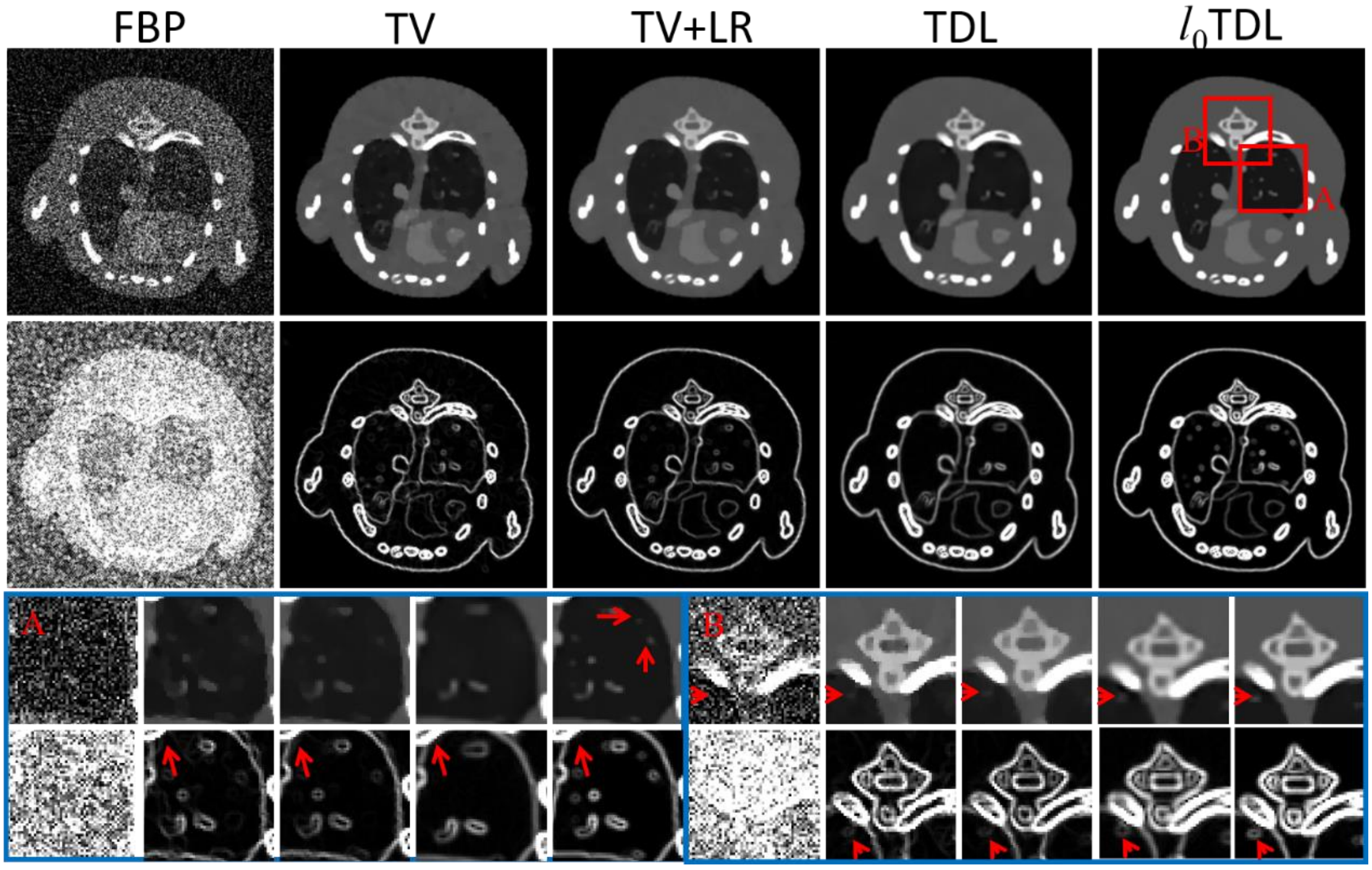

Fig. 9. Reconstruction results of the modified mouse thorax phantom. The first two rows are the reconstructed and gradient images from 160 projections and the last two rows are the magnified images of ROIs A and B. The display window of the reconstructed images is [0 3] cm$^{-1}$ and the gradient images are in [0 0.8] cm$^{-1}$.

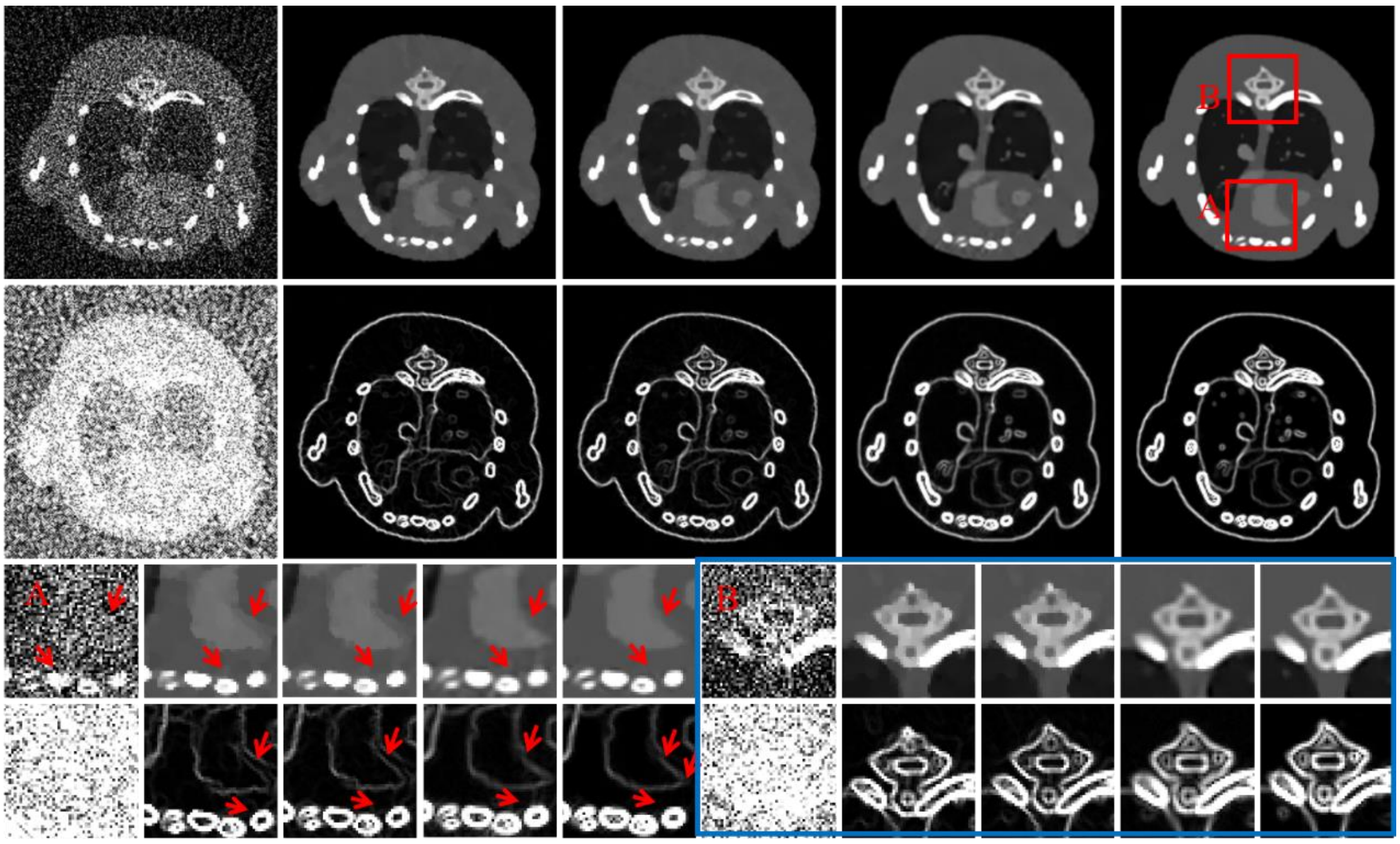

Fig. 10. Same as Fig 9 but from 106 projections.

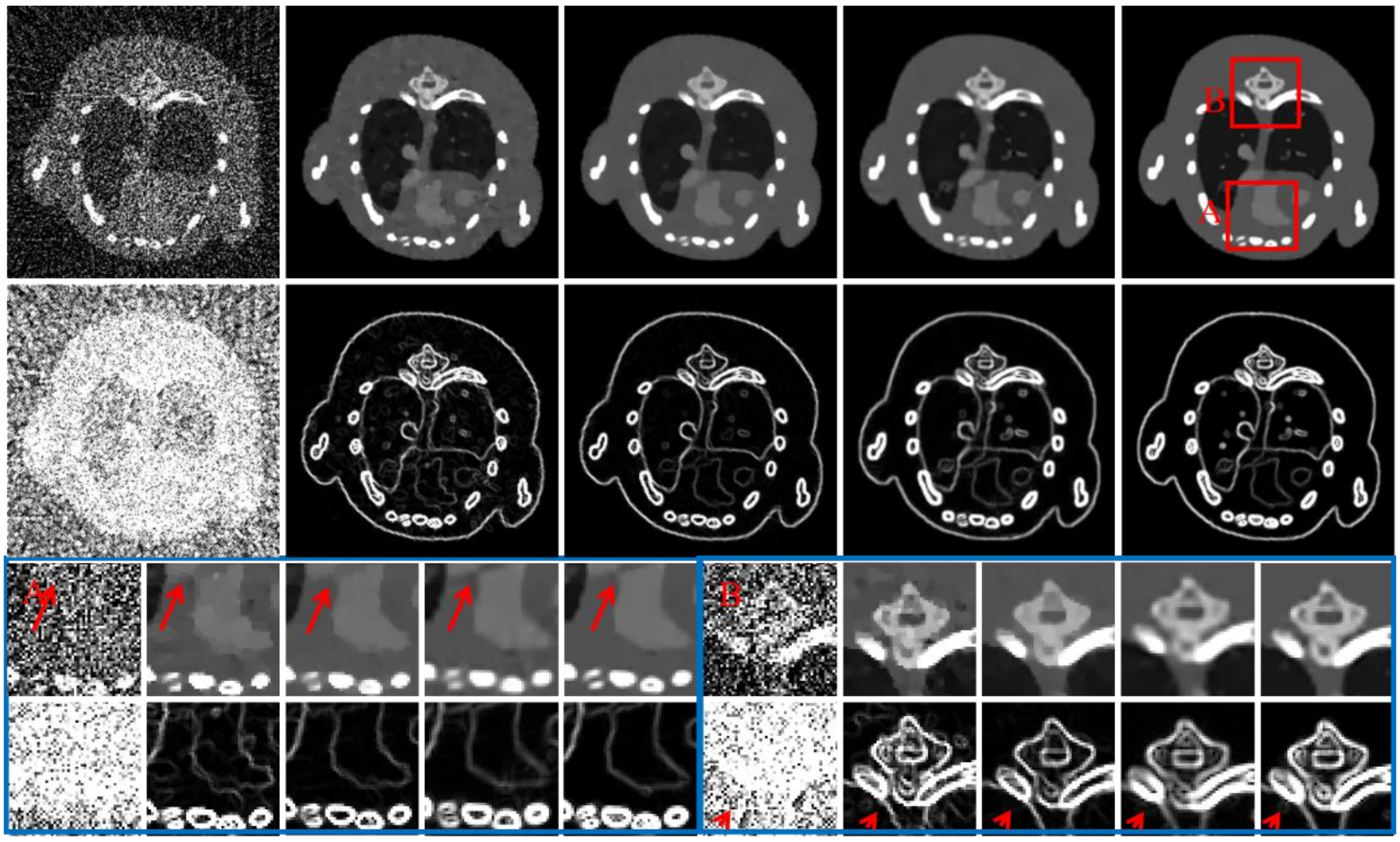

Fig. 11. Same as Fig 9 but from 80 projections.

To fairly compare the image quality for different algorithms, the parameters in the competing and proposed methods are optimized based on the RMSE minimization strategy, and the cases with minimal RMSEs are selected for further comparison. The reference image is reconstructed by the FBP algorithm from full noise-free data. From the results in Figs. 9, 10 and 11, one can see that the proposed method can obtain the best image quality compared with other competing techniques. Compared with the FBP, TV, TV+LR and TDL algorithms, our algorithm can not only remains the image edge information but also has good performance in improving the capability of anti-noising and further recovering minor structures. The magnified regions of interest (ROIs) (Figs. 9, 10 and 11) and the corresponding gradient images confirm the huge advantages of the developed $\ell_0$TDL technique. To quantitatively evaluate the performance of different techniques in sparse-view reconstruction, the results of RMSE, SSIM and FSIM are listed in table 2. The quantitative analysis results of FBP are omitted in table 2. Table 2 shows that the $\ell_0$TDL method has the smallest RMSE in all the representative channels ($1^{st}$, $4^{th}$ and $8^{th}$ channels) over all different sparse view projections, followed by the TV+LR method which has slightly smaller RMSE than the TDL for lower energy channels. The TDL algorithm slightly outperforms the TV+LR method for higher energy channels. Obviously, the TV minimization based iterative algorithm has the largest RMSE in all channels. Comparing the images from different sparse views, we can observe that the greater the number of projections is, the higher the image quality is. Especially, when the projection number equals to 160, it becomes clear for the fine soft tissues and the surrounding of bony structures. The same conclusions are made from the indexes of FSIM and SSIM, which usually measure the similarity between two images. As we can see from table 2, with the increase of energy channel index, the signal noise ratio decreases and the image quality becomes worse.

Table 2. Quantitative evaluation of different projection views reconstruction results.

| Views | | RMSE | | | SSIM | | | FSIM | | |
|---|---|---|---|---|---|---|---|---|---|---|
| | Channel | $1^{st}$ | $4^{th}$ | $8^{th}$ | $1^{st}$ | $4^{th}$ | $8^{th}$ | $1^{st}$ | $4^{th}$ | $8^{th}$ |
| 80 | TV | 0.1975 | 0.0677 | 0.0346 | 0.9037 | 0.8988 | 0.8472 | 0.8993 | 0.8908 | 0.8464 |
| | TV+LR | 0.1828 | 0.0633 | 0.0319 | 0.9247 | 0.9184 | 0.8882 | 0.9141 | 0.9102 | 0.8850 |
| | TDL | 0.1854 | 0.0638 | 0.0285 | 0.9246 | 0.9164 | 0.8893 | 0.9019 | 0.9044 | 0.8770 |
| | $\ell_0$TDL | **0.1757** | **0.0603** | **0.0269** | **0.9357** | **0.9305** | **0.9050** | **0.9247** | **0.9209** | **0.8957** |
| 106 | TV | 0.1912 | 0.0679 | 0.0331 | 0.9169 | 0.9039 | 0.8614 | 0.9079 | 0.8976 | 0.8655 |
| | TV+LR | 0.1804 | 0.0629 | 0.0330 | 0.9263 | 0.9235 | 0.8963 | 0.9158 | 0.9146 | 0.8900 |
| | TDL | 0.1800 | 0.0620 | 0.0275 | 0.9302 | 0.9229 | 0.8958 | 0.9170 | 0.9113 | 0.8849 |
| | $\ell_0$TDL | **0.1754** | **0.0587** | **0.0267** | **0.9378** | **0.9320** | **0.9087** | **0.9271** | **0.9238** | **0.8986** |
| 160 | TV | 0.1908 | 0.0661 | 0.0315 | 0.9257 | 0.9097 | 0.8736 | 0.9167 | 0.9023 | 0.8757 |
| | TV+LR | 0.1769 | 0.0617 | 0.0347 | 0.9349 | 0.9287 | 0.9003 | 0.9244 | 0.9195 | 0.8939 |
| | TDL | 0.1775 | 0.0612 | 0.0271 | 0.9330 | 0.9254 | 0.9008 | 0.9222 | 0.9163 | 0.8932 |
| | $\ell_0$TDL | **0.1751** | **0.0579** | **0.0264** | **0.9409** | **0.9360** | **0.9124** | **0.9282** | **0.9243** | **0.9031** |

The attenuation coefficients of three basis materials (soft, bone and iodine) and the relative biases are compared for 80 views. A relative bias is computed as the ratio between the absolute bias and the corresponding mean value of the reference in an energy channel. For simplicity, we demonstrate the results from 80 views with different iterative algorithms in each channel in Fig. 12. The reference mean values of different materials are generated by FBP algorithm from noisy-free projection dataset. The TV method tends to smooth small structures and results in the greatest relative bias for bone (up to 10.0% in channel 8) in sparse views reconstruction. Of course, the disadvantage remains in the TV+LR algorithm where the greatest relative bias can reach 9.2% in $8^{th}$ channel. The mean values of bone $\ell_0$TDL images are the most accurate and the relative biases are below 1.8% in all channels. The next one is the TDL algorithm. For the iodine contrast agent, because of the spectral flattening effect near the K-edge of iodine, the TV+LR has a poor performance with a 14.9% relative bias in channel 6, and the relative biases from other techniques are no more than 2.5%. Particularly, the relative biases of iodine from the $\ell_0$TDL are below 1.6%. Regarding the soft tissues, the relative bias in energy channel 8 from the TV+LR reaches 3.6%. However, the relative biases of the TV and $\ell_0$TDL methods are only below 0.8%. From the corresponding mean values of different algorithms for the soft tissue, one can see the $\ell_0$TDL has minimal values compared with other methods in lower energy channels, and it is comparable with the TV based method in higher energy channels.

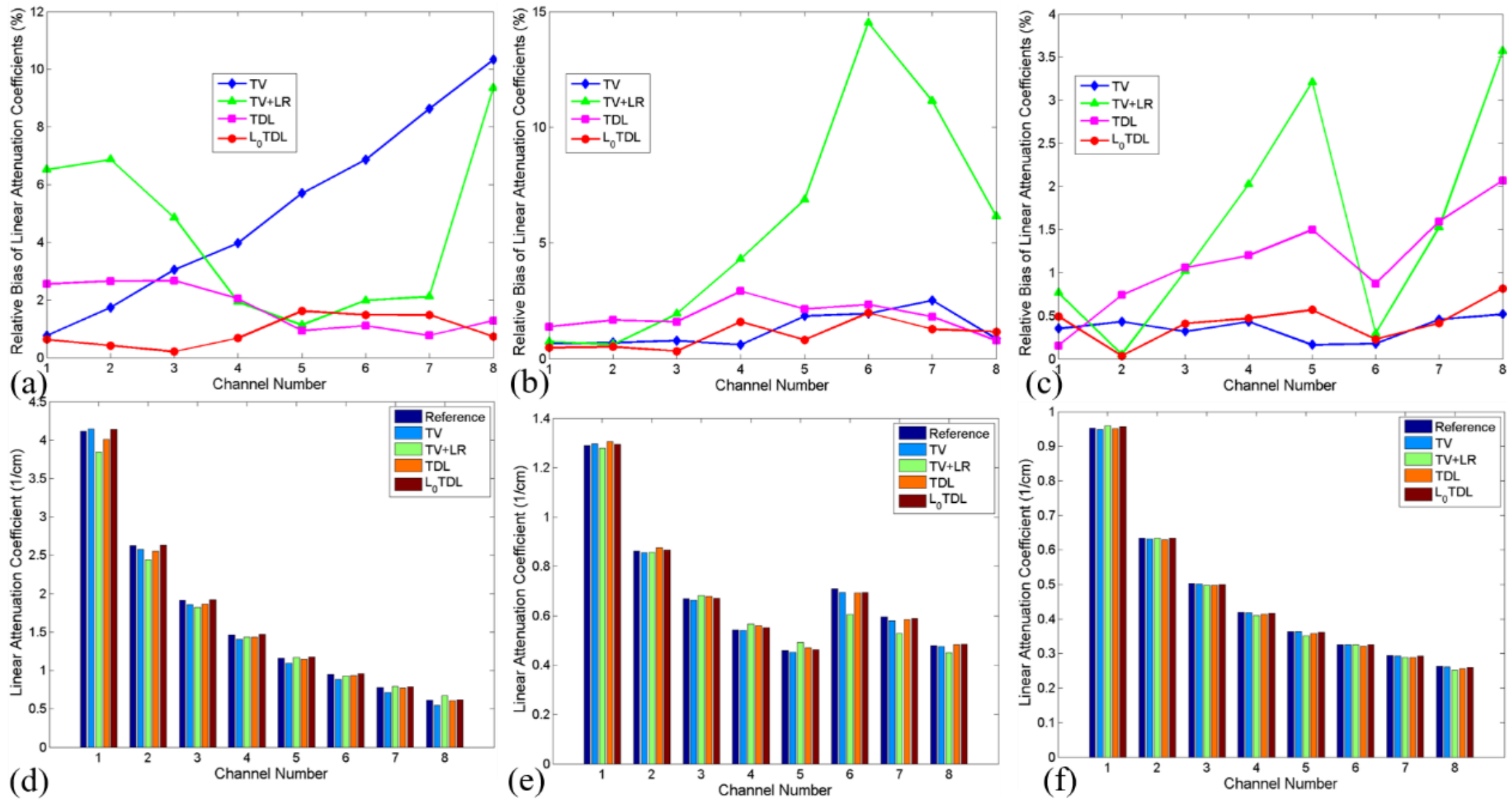

Fig. 12. Mean values and the corresponding relative biases for bone (1st column), iodine contrast agent (2nd column) and soft tissue (3rd column).

To evaluate the performance of the proposed $\ell_0$TDL method on the decomposition accuracy of three basis materials, the reconstructed spectral images from 80 projections by the TV, TV+LR, TDL and $\ell_0$TDL are decomposed into bone, soft tissue and iodine contrast agent, respectively. The decomposed results of three basis materials and the corresponding color images are shown in Fig. 13. From the first row of Fig. 13, we can see the TV, TV+LR and TDL methods wrongly classify the bony region, and the $\ell_0$TDL has a unique advantage on the most accurate bone components. For the soft-tissue component decomposition, the results with the proposed $\ell_0$TDL provide much finer structures compared with other competing techniques (2nd row in Fig. 13). As for the iodine decomposition, it seems the TDL and $\ell_0$TDL offer similar accuracy in contrast with the TV and TV+LR algorithms.

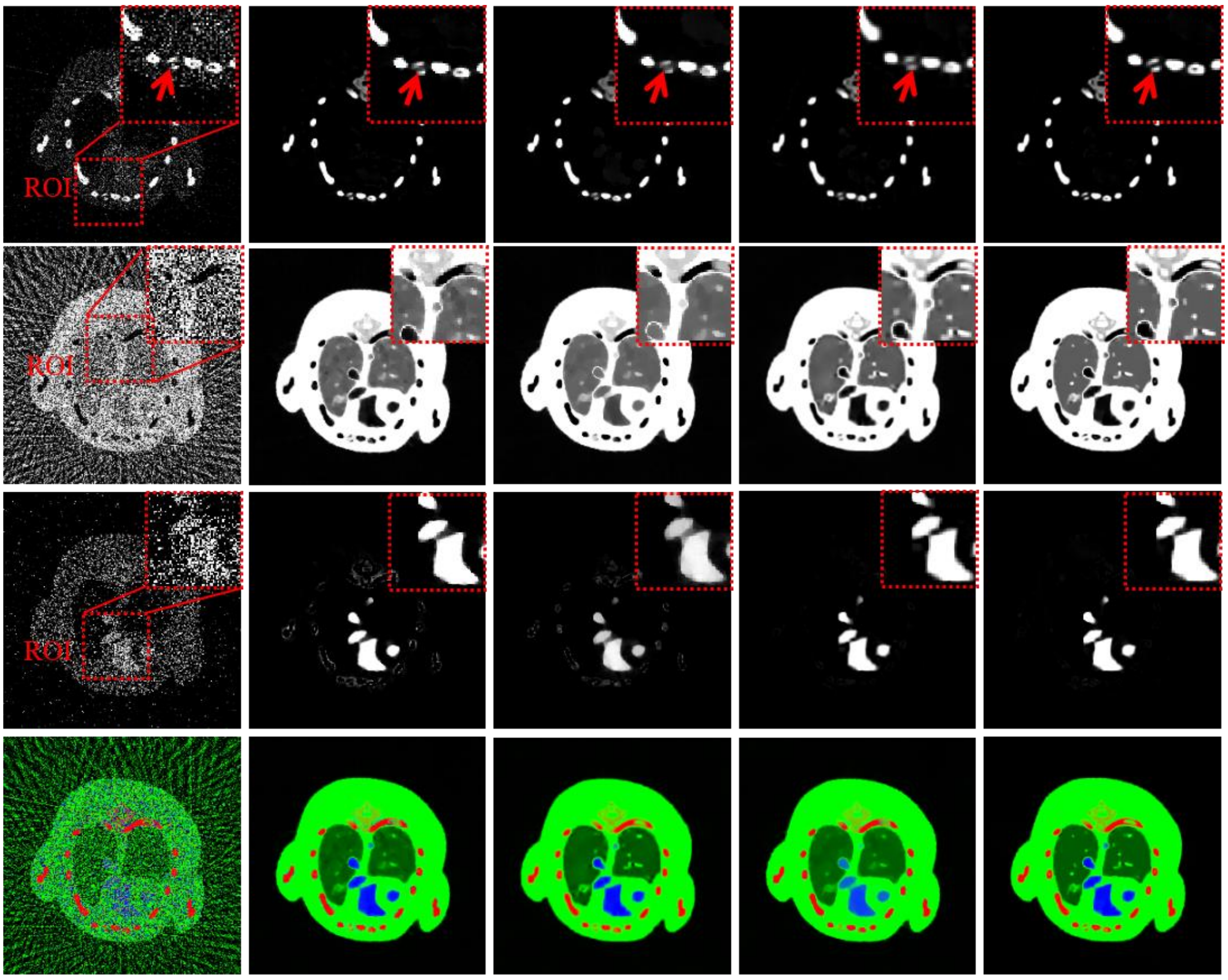

Fig. 13. Material decomposition results of images reconstructed by different algorithms from 80 projections. The 1st to 3rd rows are the decomposed bone, soft tissue and iodine contrast agent components, respectively. The 4th row is the true color images where red, green and blue regions represent the three basis materials.

In order to investigate the convergences of the proposed $\ell_0$TDL and other competing methods, the convergence curves in terms of averaged RMSEs *vs.* iteration number are given in Fig. 14. Compared with other competing algorithms, the $\ell_0$TDL method can converge to an optimized solution quickly with a smaller RMSE. From Fig. 14, for the $\ell_0$TDL method, one can see that the RMSE decreases rapidly at first and then it is subsequently stable after 40 iterations.

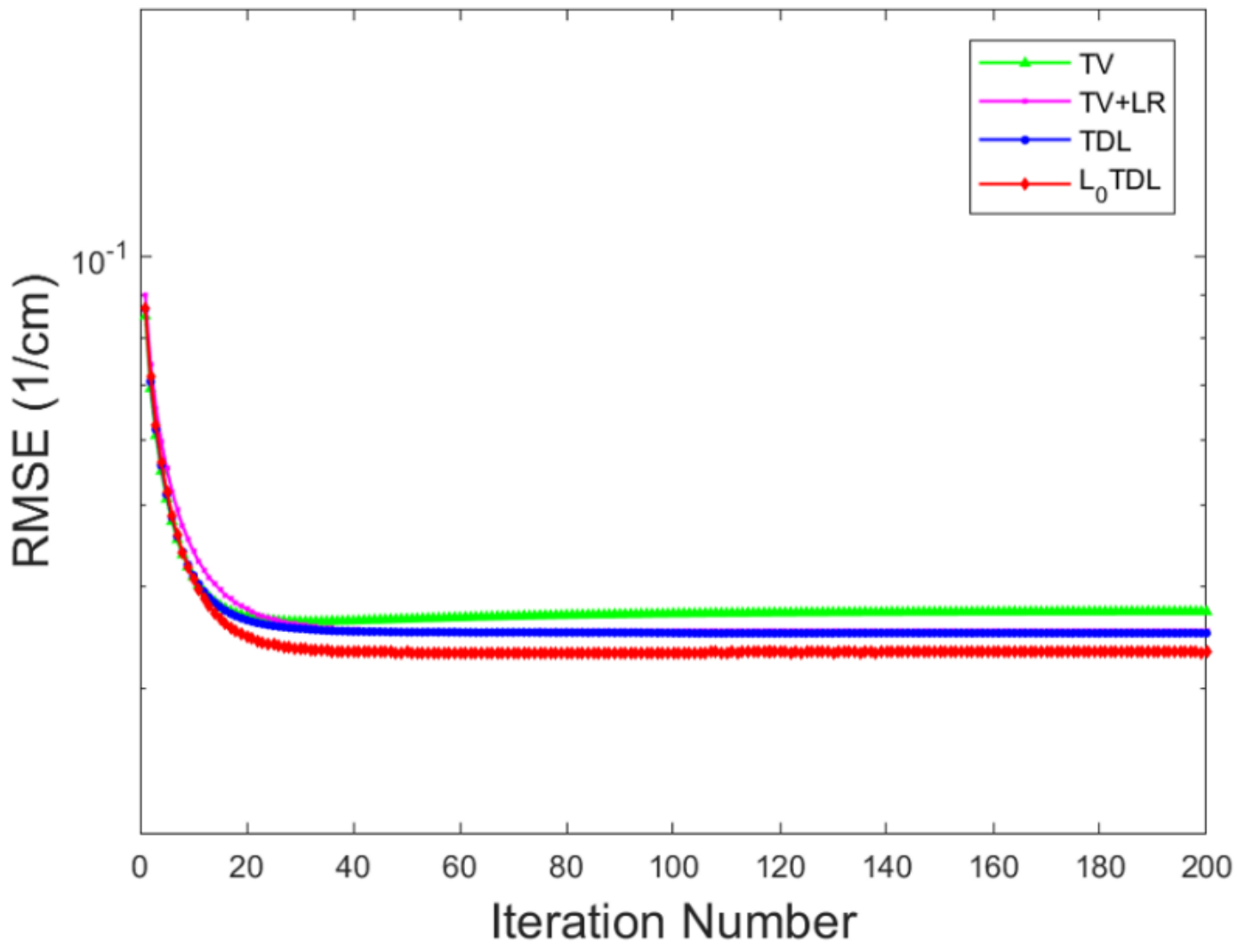

Fig. 14.The convergence curves in terms of average RMSE *vs.* iteration number.

All the reconstruction algorithms are implemented under the same condition. For the case of 80 projections, the TV, TV+LR, TDL and $\ell_0$TDL methods take 4.60, 4.70, 5.54, 10.68 seconds per iteration, respectively. The proposed $\ell_0$TDL optimization needs longer time than other competing methods due to the image gradient $\ell_0$–norm minimization. Given the fact that the image gradient $\ell_0$–norm optimization process is implemented channel by channel, this process can be implemented in parallel on GPU.

*4.A.2 Low-dose reconstruction*

To evaluate the performance of the proposed $\ell_0$TDL algorithm for low-dose reconstruction, we generate low-dose projection datasets by reducing the photon number of each x-ray path. The photon number from the source are set as $4\times10^3$ and $3\times10^3$ to simulate different low-dose levels, respectively. Because the photon number of each x-ray is reduced, the projection dataset will be further smeared by Poisson noise. Fig. 15 presents different channel images reconstructed from low-dose projection datasets with 80 views using different methods. From Fig. 15, we can see that our proposed $\ell_0$TDL method can provide much finer structures compared with other methods.

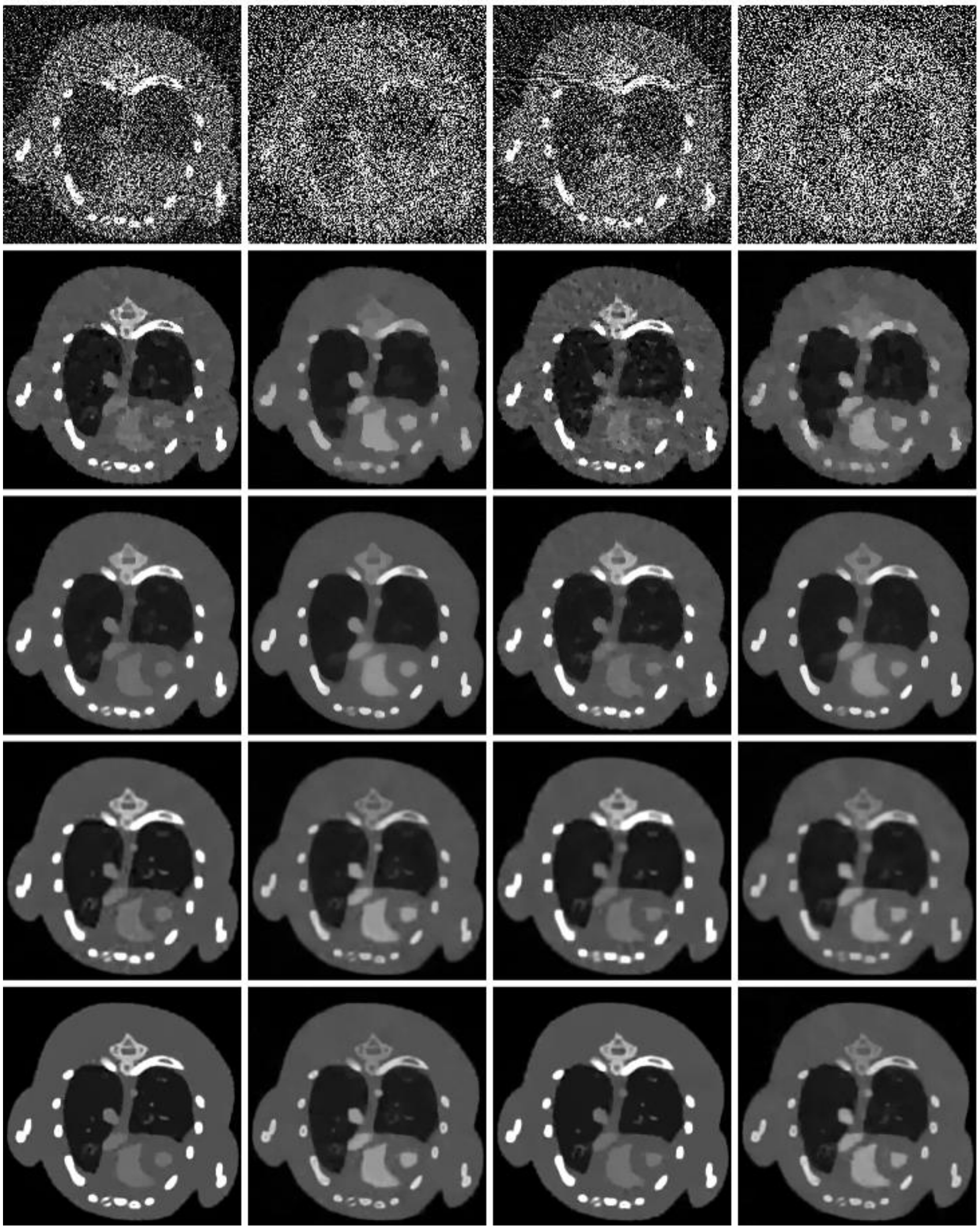

Fig. 15. Reconstructed images from low-dose projections. The 1st and 3rd columns are channel 1 images in a display window [0 3] $cm^{-1}$, and the 2nd and 4th columns are channel 8 images in a display window [0 0.8] $cm^{-1}$. The first two columns are reconstructed from datasets with $4x10^3$ photons, and the last two columns are reconstructed from datasets with $3x10^3$ photons. From the 1st to 5th rows, the images are reconstructed by the FBP, TV, TV+LR, TDL and the proposed $\ell_0$TDL algorithms, respectively.

To quantitatively analyze the performance of our proposed algorithm for low-dose reconstruction, the averages of RMSE for different channel are given in Table 3. For the photon number $4\times10^3$, the average RMSE of different channels using the analytic FBP method is greater than other iterative techniques. The TV-based regularization algorithm has a maximum RMSE value, followed by the TDL and TV+LR. Of course, our proposed $\ell_0$TDL can reach a minimal RMSE among all the competing methods. In terms of SSIM and FSIM, the $\ell_0$TDL method always has the maximal values compared with other methods.

Table 3. The quantitative assessment of reconstructed image quality with low-dose projection data (unit: $10^{-2}$).

| | | FBP | TV | TV+LR | TDL | $\ell_0$TDL |
|---|---|---|---|---|---|---|
| $4\times10^3$ | RMSE | 56.02 | 9.77 | 9.09 | 9.13 | **8.68** |
| | SSIM | 34.10 | 88.33 | 90.71 | 90.86 | **92.20** |
| | FSIM | 48.42 | 87.99 | 90.08 | 89.46 | **91.27** |
| $3\times10^3$ | RMSE | 63.91 | 10.22 | 9.23 | 9.68 | **8.74** |
| | SSIM | 33.00 | 85.89 | 90.38 | 89.75 | **91.90** |
| | FSIM | 47.15 | 86.11 | 89.78 | 88.43 | **90.96** |

To test the performance of our proposed algorithm for accuracy of basis material decomposition, the RMSEs between the reference image, which is reconstructed by FBP algorithm with normal dose and noisy-free projection data, and reconstructed image from low-dose projection reconstruction with all iterative methods were given in Table 4. From the results, we can see the $\ell_0$TDL algorithm always obtains the minimal RMSEs of all basis materials.

Table 4. The RMSEs of different decomposed components (unit: $10^{-4}$) with low-dose datasets.

| | | TV | TV+LR | TDL | $\ell_0$TDL |
|---|---|---|---|---|---|
| $4\times10^3$ | Soft tissue | 2.9898 | 3.3274 | 2.8581 | **2.7486** |
| | Iodine | 1.6665 | 2.2706 | 1.2139 | **1.1220** |
| | Bone | 1.6819 | 1.6335 | 1.6732 | **1.5881** |
| $3\times10^3$ | Soft tissue | 3.0963 | 3.3647 | 3.0116 | **2.7806** |
| | Iodine | 1.7836 | 2.3120 | 1.3398 | **1.1974** |
| | Bone | 1.6929 | 1.6671 | 1.7752 | **1.5993** |

### *4.B. Realistic Mouse dataset reconstruction*

To demonstrate the advantages of the proposed ℓ0TDL algorithm for low-dose spectral reconstruction in practical applications, an injected gold nanoparticles (GNP) mouse is scanned by a CT system including one x-ray source and one photon counting detector. In this system, the distance from the x-ray source to the PCD is set as 255 mm, the distance between the x-ray source and rotation axis is 158 mm, and 371 projections are uniformly acquired over a full scan circular trajectory. The energy spectrum of x-ray source is divided into 13 channels via multiple scans, i.e., 13 images can be reconstructed for one slice. The PCD consists of 512 elements, each of which covers a length of 0.11 mm, and the radius of FOV is 9.21mm. The detector offset for this datasets was 1.0 mm. Fig. 16 shows some representative channel images reconstructed by the FBP algorithm from full projections, where each channel image is a matrix of 512×512 covering an area of 18.41×18.41mm$^2$. From Fig. 16, we can see the reconstructed images include severe artifacts, implying that the real dataset is severely tainted by noise and the result in the dataset has a low signal noise ratio (SNR). This case can be considered as the low dose real projection dataset. Thus, we only investigate the sparse view reconstruction for this real dataset.

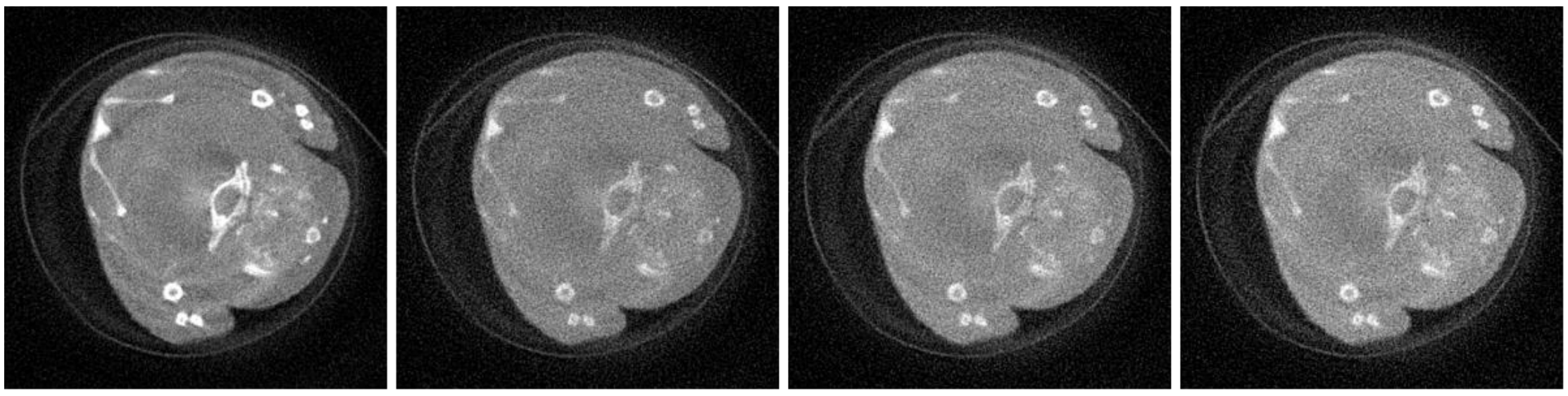

Fig.16. From the left to right columns, images are reconstructed for the 1$^{st}$, 4$^{th}$, 9$^{th}$ and 13$^{th}$ channels and the display window is [0, 0.8] cm$^{-1}$.

To validate the performance of proposed the $\ell_0$TDL method on sparse views reconstruction, Fig. 17 demonstrates the results from 120 views of the channel 1. From the two extracted ROIs A and B, we can observe the proposed method has the best ability of protecting image edge. From the magnified ROI B, we can see the x-ray beam-hardening artifacts can be further reduced using our method. The image gradient $\ell_0$–norm can penalize the sparsity in image gradient domain, which may result in reduced beam-hardening artifacts. Fig. 18 shows three basic material decomposition of Fig. 17. From Fig. 18, one can see the most accuracy material decomposition can be obtained by the proposed algorithm compared with other competing techniques. In terms of the rendered color image, the image edge of the $\ell_0$TDL method is more clearly than other algorithms.

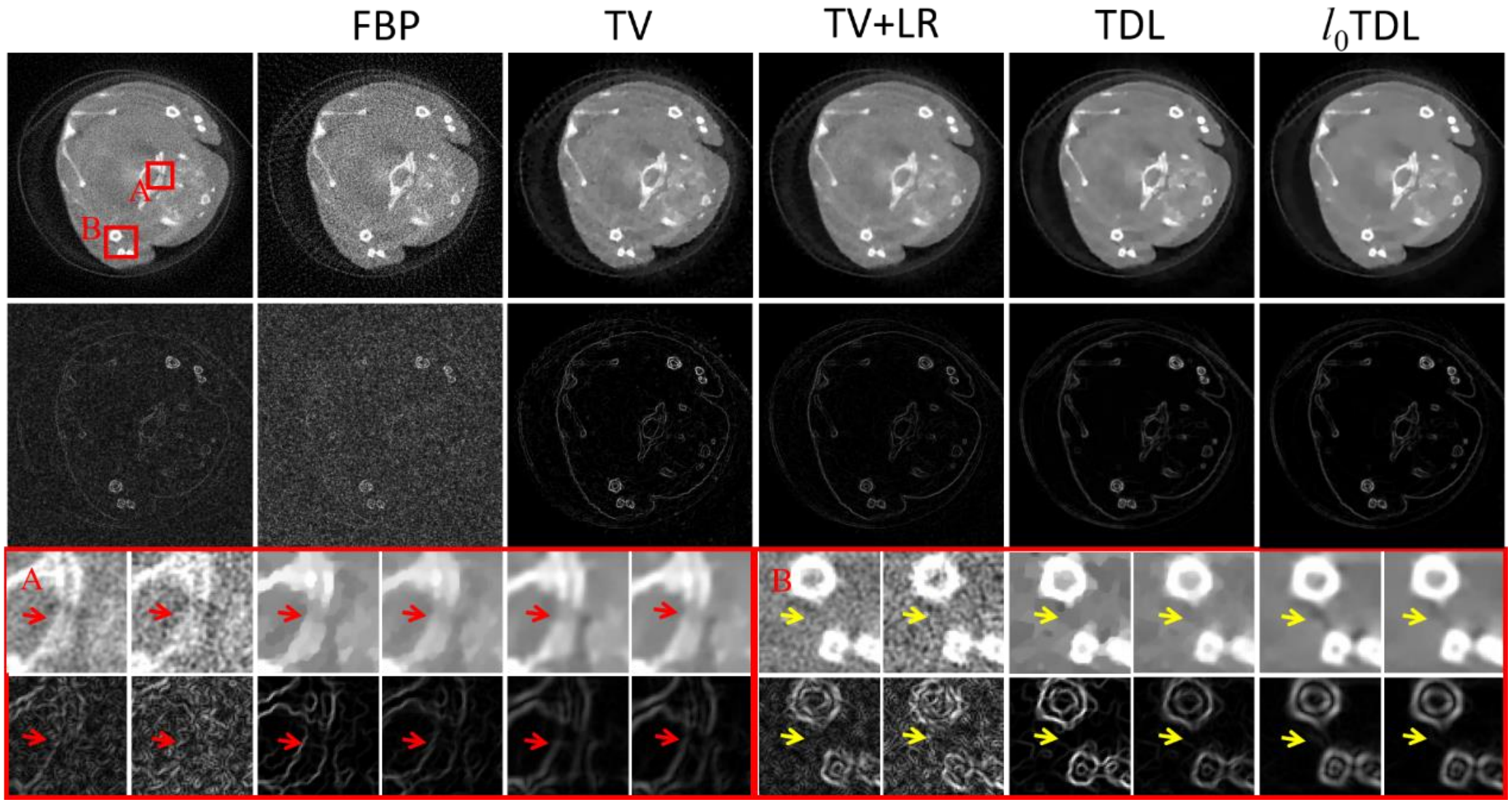

Fig. 17. Same as Fig. 9 but reconstructed from the 120 projections of realistic mouse dataset. The first column is the originally reconstructed image using the FBP algorithm from full projections. The display window of the reconstructed images is [0, 0.8] cm$^{-1}$ and the gradient images are in [0, 0.4] cm$^{-1}$.

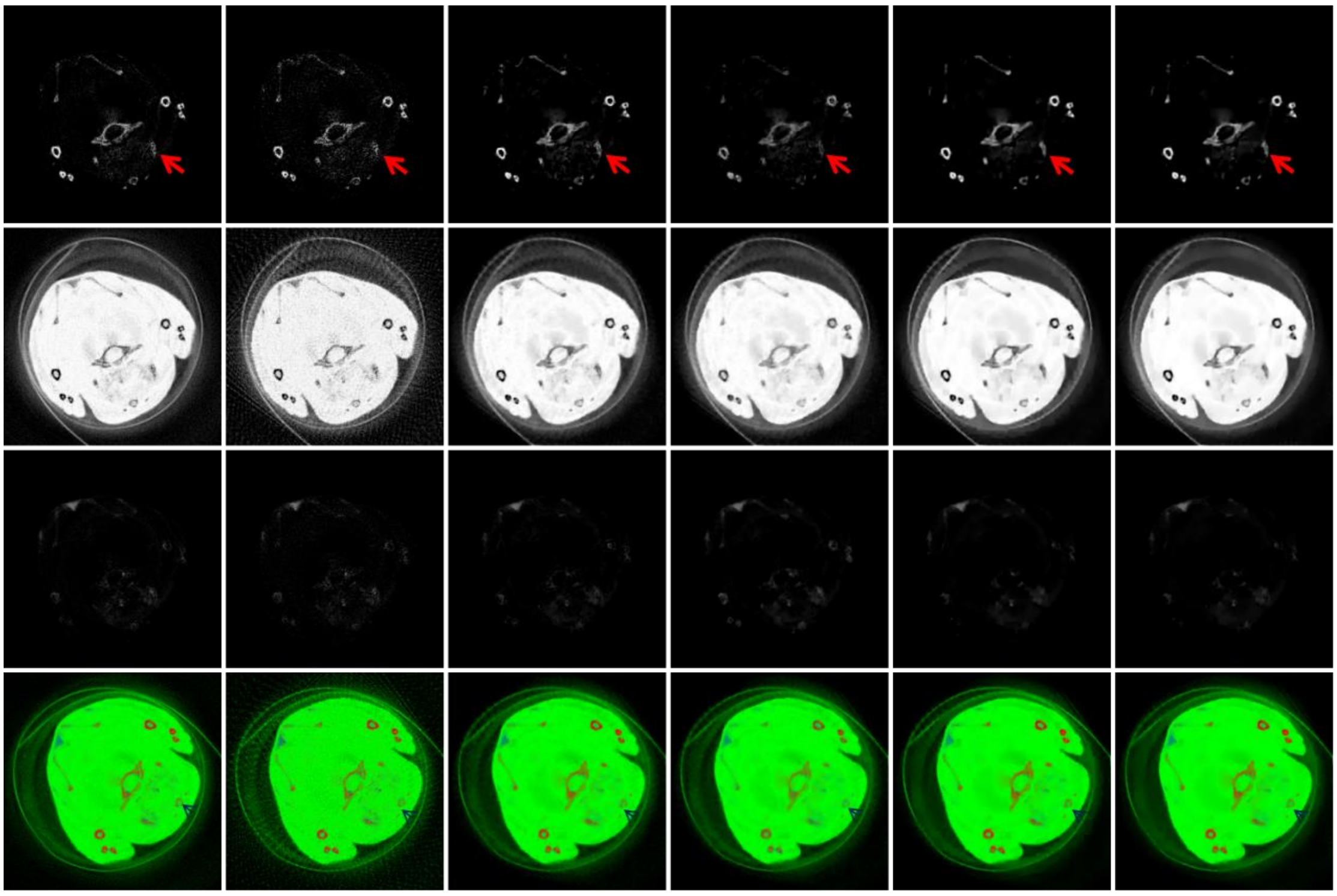

Fig. 18. The three basic material decomposition of Fig.17. From the first to third rows present the decomposition of bone, soft tissue and GNP. The fourth row is the fusion color image, where the red, green and blue represent the bone, soft tissue and GNP respectively.

Fig. 19 shows the relative results of 80 projections from different methods. From the first row, we can observe that the images reconstructed by the FBP and TV methods contain severe artifacts. The TV-based method can induce staircase artifacts in the reconstructed image and further smeared some finer structures. The TV+LR method can not distinguish the structures indicated by the red and yellow arrows. For the TDL method, the region indicated by the yellow arrow is polluted by severe artifacts. It is very difficult to distinguish the bone edge. As indicated by the red and yellow arrows, the edge information is well protected in the reconstructed image from the $\ell_0$TDL method.

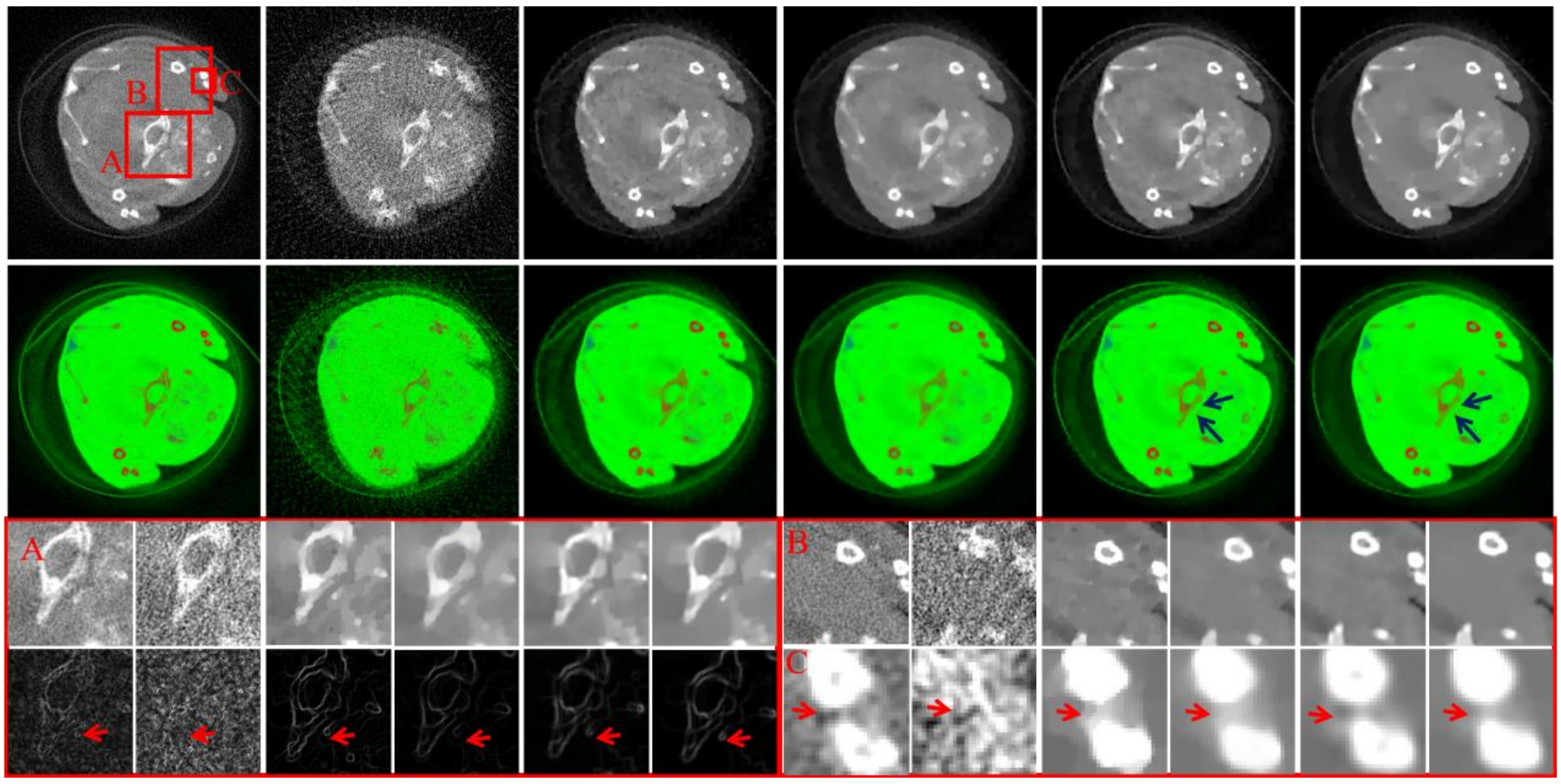

Fig. 19. Same as the Fig.17 but reconstructed from the 80 projections. The second row images are color images instead of gradient images.

To further investigate the behavior of the $\ell_0$TDL algorithm in sparse views reconstruction, the reconstructed results from only 40 projections using different methods are given in Fig. 20. From the magnified ROIs A and B, we can see the $\ell_0$TDL has a great potential in preserving image edge, which is confirmed by bone boundaries indicated by red arrows. From the viewpoint of material decomposition, the TV+LR method can make wrongly bone material decomposition. The edge of boneusing the $\ell_0$TDL method is protected very well compared with other iteration algorithms.

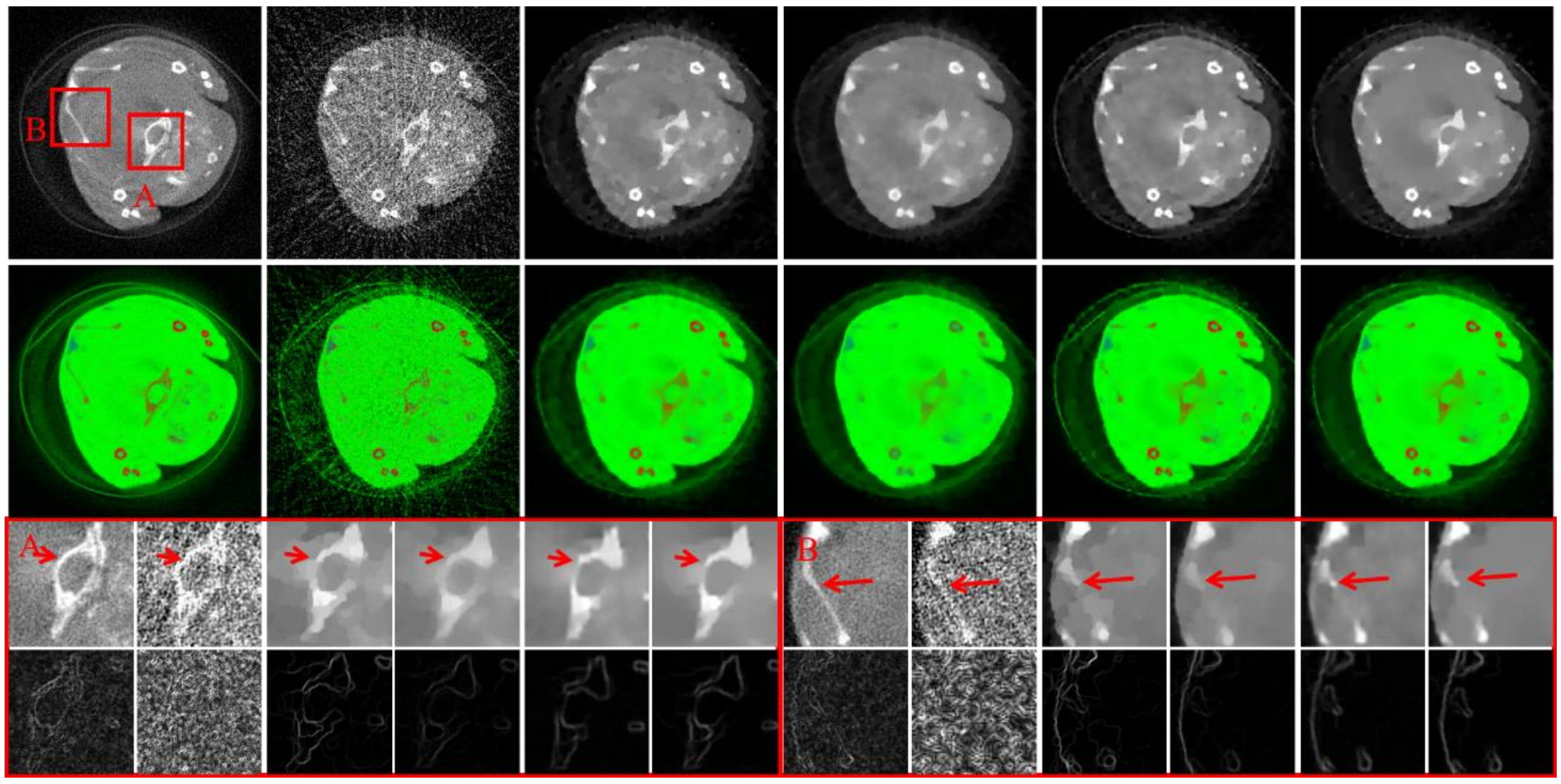

Fig. 20. Same as the Fig.19 but reconstructed from 40 projections.

# 5. Discussion and conclusions

To penalize the image gradient for preserving image edge information from each channel and improve the anti-noising capability of the TDL method, we develop an $\ell_0$TDL algorithm for low-dose spectral CT reconstruction. By incorporating the image gradient $\ell_0$–norm into the TDL based reconstruction framework, the image quality of channel reconstructions is dramatically improved, especially in the cases of low-dose and sparse-view reconstruction. Both numerical simulations and realistic preclinical mouse study confirm that the proposed $\ell_0$TDL algorithm outperforms the TV, TV+LR, and TDL methods.

The fine image structure and sharp image edge provide relatively greater image gradient magnitude, which can be easily smoothed during the course of the total variation minimization. However, the number of non-zero components of the sharp image edge is always a constant. Because the image gradient $\ell0$–norm only concentrates on calculating the number of non-zero values contained in reconstructed gradient image, it implies that the image gradient $\ell0$–norm can protect image edges and keep fine structures. The $\ell_0$TDL algorithm also has advantages in suppressing ring artifacts. Fig. 21 shows the full-projection-based images by the TDL and $\ell_0$TDL approaches. The artifacts (indicated by the red arrows) can be reduced by our algorithm around the bone region. We also show the difference images reconstructed from different number of projections (Fig. 21), where a full-projection-based reconstruction image by the $\ell_0$TDL is chosen as reference. Comparing the difference images, we can see the outstanding performance of our proposed algorithm in sparse view reconstruction.

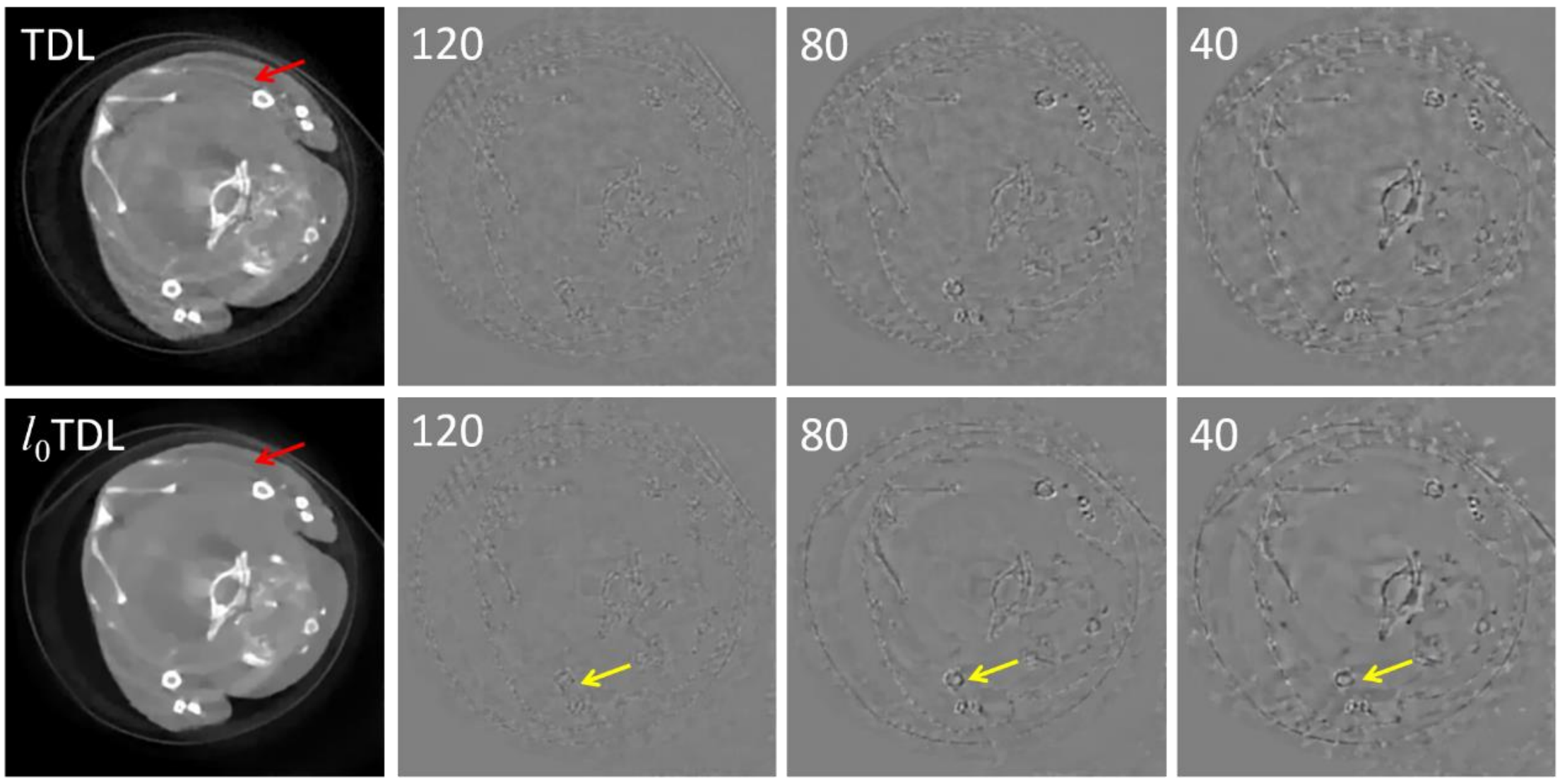

Fig. 21. Reconstructed images and the corresponding difference images for 1st channel with respect to different views using the TDL and $\ell_0$TDL methods. The display windows of the reconstructed images and difference images are [0, 0.8] $cm^{-1}$ and [-0.3, 0.3] $cm^{-1}$, respectively.

A natural question is that if the proposed $\ell_0$TDL outperform the TV based TDL method. That is, what will happen if we relax the $\ell_0$–norm to $\ell_1$–norm for the image gradients. To further explore the advantages of the $\ell_0$TDL in preserving image edges, we also implement the TV+TDL and compare it with our $\ell_0$TDL algorithm. Fig. 22 shows the results for the case of 80 views of numerical simulations. From Fig. 22, we can easily find that the $\ell_0$TDL method can better recover fine structures and protect image edge information than the TV+TDL method. Meanwhile, the realistic experiments demonstrate consistent conclusions. Note that the number of non-zero components of each energy-channel is almost equal. Compared with the TV based TDL method, the regularized parameter of the $\ell_0$-norm term in the $\ell_0$TDL can be set as a uniform value.

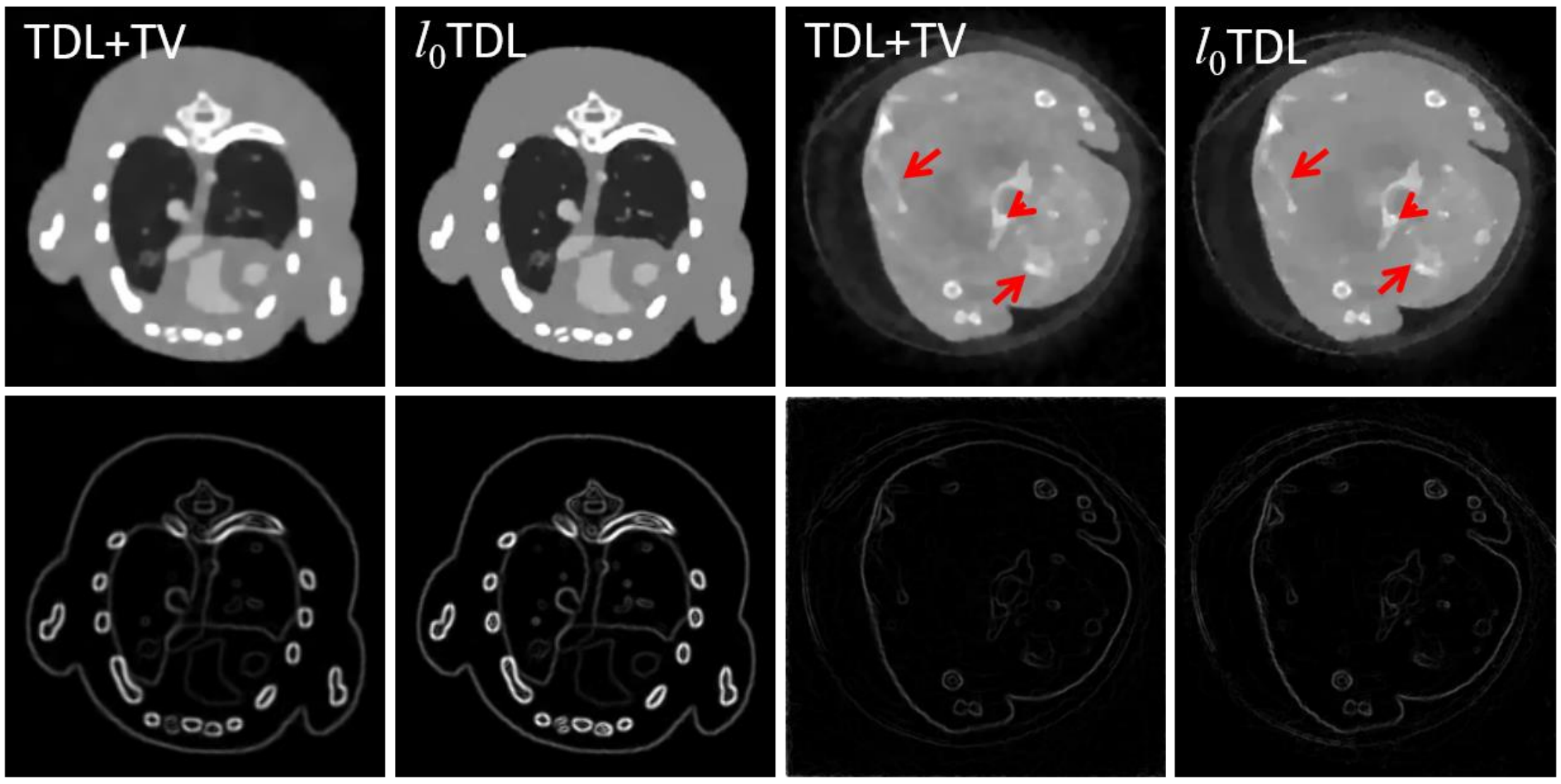

Fig. 22. The first two columns represent the reconstructed and corresponding gradient images from the 4th channel of numerically simulated mouse dataset using the TDL and $\ell_0$TDL methods. The display window of the reconstructed image is [0, 0.8] $cm^{-1}$. The last two columns are the same as the first two columns but from the 13th channel of the realistic dataset and the display window of gradient images is [0, 0.4] $cm^{-1}$.

Although exciting results have been achieved by using the developed $\ell_0$TDL method, there are still some issues. First, numerous parameters in the $\ell_0$TDL method, including the regularization and control parameters, need to be optimized. In this work, we select the parameters empirically based on extensive experiments. This indicates the parameter influence in light of image quality assessment. However, the theoretical analyses and optimizations are still open problems that need to be further investigated. In addition, in our proposed $\ell_0$TDL algorithm, the global tensor dictionary is trained from prior image reconstructed by the FBP from full projections. However, such a global tensor dictionary is not available in some cases. For those circumstances, we have to utilize sparse-view projection based reconstruction image or other similar images to train the tensor dictionary, which may compromise the image quality. In this case, we may employ the adaptive dictionary learning technique to train and update the tensor dictionary during the iteration process from low-dose datasets with more computational cost.

In conclusion, we propose an $\ell_0$TDL algorithm based on a global tensor dictionary and image gradient $\ell_0$ for low-dose spectral CT reconstruction. The developed $\ell_0$TDL method can not only well maintain fine structures and image edges, but also reduce beam-hardening artifacts especially in the areas of bone. This will be extremely meaningful for low-dose spectral CT reconstruction.

# Acknowledgement

This work was supported by National Natural Science Foundation of China (No. 61471070), National Instrumentation Program of China (No. 2013YQ030629), NIH/NIBIB U01 grant (EB017140) and China Scholarship Council (No. 201706050070). The authors would like to thank the MARS team in New Zealand for the realistic mouse datasets. The authors are grateful to Mr. Changcheng Gong at Chongqing University for his suggestions. The authors thank Mr. Qi Li for his extensive discussions and valuable suggestions. The authors are also grateful to the anonymous reviewers for their valuable comments.